\documentclass[twoside,11pt]{article}
\usepackage{pdfpages}
\usepackage{graphicx}
\graphicspath{ {images/} }
\usepackage{dsfont}
\usepackage{bm}
\usepackage{url}            
\usepackage{booktabs}       
\usepackage{amsfonts}       
\usepackage{nicefrac}       
\usepackage{microtype}      
\usepackage{amssymb}
\usepackage{amsmath}
\usepackage{dsfont}
\usepackage{multirow}
\usepackage{a4wide}
\usepackage{natbib}
\usepackage[]{algorithm2e}
\usepackage{afterpage}
\usepackage{float}
\usepackage{hyperref}
\hypersetup{
    colorlinks=true,
    linkcolor=blue,
    filecolor=magenta,
    urlcolor=cyan,
}

\begin{document}

\title{Dealing with Categorical and Integer-valued Variables in Bayesian Optimization with Gaussian Processes}

\author{%
  Eduardo C. Garrido-Merch\'an \\
  eduardo.garrido@uam.es \\
  Universidad Aut\'onoma de Madrid\\
  Calle Francisco Tom\'as y Valiente 11,  \\
  Madrid, 28049, Spain
  \and
  Daniel Hern\'andez-Lobato \\
  daniel.hernandez@uam.es \\
  Universidad Aut\'onoma de Madrid\\
  Calle Francisco Tom\'as y Valiente 11,  \\
  Madrid, 28049, Spain
}

\date{}

\maketitle

\maketitle

\begin{abstract}
Bayesian Optimization (BO) is useful for optimizing functions
that are expensive to evaluate, lack an analytical expression and whose 
evaluations can be contaminated by noise. These methods rely on a probabilistic model of the 
objective function, typically a Gaussian process (GP), upon which an acquisition function is built.
The acquisition function guides the optimization process and measures the expected utility of 
performing an evaluation of the objective at a new point. GPs assume continuous input variables. 
When this is not the case, for example when some of the input variables take categorical or integer 
values, one has to introduce extra approximations. 
Consider a suggested input location taking values in the real line.
Before doing the evaluation of the objective, a common approach 
is to use a one hot encoding approximation for categorical variables, or to round to the closest 
integer, in the case of integer-valued variables. 
We show that this can lead to optimization problems 
and describe a more principled approach to account for input 
variables that are categorical or integer-valued. We illustrate in both synthetic and a real 
experiments the utility of our approach, which significantly improves the results of
standard BO methods using Gaussian processes on problems with categorical or integer-valued 
variables.
\end{abstract}

\section{Introduction} \label{sec:background_bo}

Many problems involve the optimization of a function with no analytical form. 
An example is tuning the parameters of the control system of a robot to maximize locomotion
speed \citep{lizotte2007automatic}. There is no closed-form expression to describe the 
function that, given specific values for these parameters, returns an estimate of the corresponding 
speed. A practical experiment with the robot or a computer simulation will 
have to be carried out for this purpose. Moreover, the time required for such an evaluation
can be high, which means that in practice one can only perform a few evaluations of the objective. 
Importantly, these evaluations may be noisy and hence different 
for the same input parameters. The noise can simply be related to the 
environmental conditions in which the robot's experiment is performed. 
When a function has the characteristics described it is called a black-box function.
Examples of problems involving the optimization of black-box functions include automatic 
tuning machine learning hyper-parameters \citep{snoek2012}, finding optimal control parameters in 
robotics \citep{lizotte2007automatic}, optimal weather sensor placement \citep{garnett2010bayesian} or 
the optimization search strategies \citep{cornejo2018bayesian}.

Bayesian Optimization (BO) methods are popular for optimizing black-box functions with 
the characteristics described \citep{movckus1975bayesian}. More formally, BO methods optimize a real-valued 
function $f(\mathbf{x})$ over some bounded domain $\mathcal{X}$ \citep{shahriari2016}. 
The objective function is assumed to lack an analytical expression (which prevents any gradient computation),
to be very expensive to evaluate, and the evaluations are assumed to be noisy (\emph{i.e.}, rather than
observing $f(\mathbf{x})$ we observe $y = f(\mathbf{x}) + \epsilon$, with $\epsilon$ some additive noise). 
The goal of BO methods is to reduce the number of objective evaluations that need to be performed
to solve the optimization problem. For this, they iteratively suggest, in a careful and intelligent 
way, an input location in which the objective that is being optimized should be evaluated each time. 
For this, at each iteration $N=1,2,3,\ldots$ of the optimization process, BO methods fit a probabilistic 
model, typically a Gaussian process (GP) \citep{rasmussen2005book}, to the collected observations of the 
objective $\{y_i\}_{i=1}^{N-1}$. The uncertainty about the potential values of the objective 
are provided by the predictive distribution of the GP. This uncertainty is used to generate an acquisition 
function $\alpha(\cdot)$, whose value at each input location indicates the expected utility 
of evaluating $f(\cdot)$ there. The next point $\mathbf{x}_N$ at which to evaluate $f(\cdot)$ is the 
one that maximizes $\alpha(\cdot)$. After collecting this observation, the process 
is repeated. When enough data has been collected, the GP predictive mean value 
for $f(\cdot)$ can be optimized to find the solution of the problem.

The key to BO success is that evaluating the acquisition function $\alpha(\cdot)$ is very cheap compared
to the evaluation of the objective $f(\cdot)$. This is so because the acquisition function only depends on 
the GP predictive distribution for $f(\cdot)$ at a candidate point $\mathbf{x}$. Thus, $\alpha(\cdot)$ can be 
maximized with very little cost. BO methods hence spend a small amount of time thinking very carefully where 
to evaluate next the objective function with the aim of finding its optimum with 
the smallest number of evaluations. This is a useful strategy when the objective function is very expensive 
to evaluate and it can save a lot of computational time. 

A problem, however, of GPs is that these probabilistic models assume that the input variables take 
real-values. If this is not the case and, for example, some of the variables  can take categorical or integer 
values, extra approximations in the BO method have to be introduced to address this issue. In the case of 
integer-valued variables, the approximations often involve simply doing some rounding to the closest integer 
after optimizing the acquisition function. In the case of a categorical variable, one simply uses a one-hot 
encoding. This involves adding as many extra input variables as different categories this variable can take. 
Then, after optimizing the acquisition function, the extra variable that is larger is set equal to one and all 
the others equal to zero. This is the approach followed, for example, in the popular software for BO Spearmint.

We show here that the approaches described for handling categorical and integer-valued variables may
make the BO method fail. These problems can be overcome by doing the rounding (to the closest integer or 
the corresponding one-hot encoding) inside the wrapper that evaluates the objective.
Nevertheless, this will make the objective constant in some regions of the input space, \emph{i.e.}, those
rounded to the same integer value, in the case of integer-valued variables, or those that lead to the same 
one-hot encoding, in the case of categorical variables. This constant behavior of the objective will be 
ignored by the GP model. To overcome this, we introduce a transformation of the input 
variables that will lead to an alternative covariance function for the GP model. 
With this covariance function, the GP will correctly describe the objective as constant 
in particular regions of the input space, leading to better modeling results and, 
in consequence, to better optimization results.

Practical examples of optimization problems involving a mix between real, categorical and integer-valued 
variables include finding the optimal hyper-parameters of a machine learning system \citep{snoek2012}. 
Specifically, in a deep neural network, we may want to adjust the learning 
rate, the number of layers and the activation function. These two last variables can only take integer 
and categorical values, respectively, while the learning rate can take real values.
Similarly, in a gradient boosting ensemble of decision trees \citep{friedman2001greedy} we may try to 
adjust the learning rate and the maximum depth of the trees, which can only take integer values. 
Our experiments show that the proposed approach for dealing with a mix of real, categorical, and 
integer-valued variables in BO methods lead to improved results over standard techniques
and other alternatives from the literature.

The rest of the paper is organized as follows: Section \ref{sec:back} gives a 
short introduction to BO and Gaussian processes. 
Section \ref{sec:dealing} describes the proposed approach 
to deal with categorical and integer-valued variables in BO methods using GPs. Section \ref{sec:related} reviews
related methods from the literature. Section \ref{sec:experiments} describes synthetic and real-world 
experiments that show that the proposed approach has advantages over standard methods for BO and related
techniques. Finally, Section \ref{sec:conclusions} gives the conclusions of this work.

\section{Background on Gaussian Processes and Bayesian Optimization} 
\label{sec:back}

BO methods relay on a probabilistic model for the black-box function being optimized. This 
model must generate a predictive distribution for the potential values of the objective at 
each point of the input space. This predictive distribution is used to guide the search, 
by focusing only on regions of the input space that are expected to deliver the most
information about the solution of the optimization problem. Most commonly used models are Gaussian processes (GPs) 
\citep{rasmussen2005book}, Random Forests \citep{thornton2013auto}, T-Student processes \citep{shah2014student} or Deep Neural Networks 
\citep{snoek2015scalable}. In this work, we will focus on the use of GP, but the same ideas can be 
implemented in a T-Student process.

A GP is defined as a prior distribution over functions. When using a GP as the underlying model, the assumption 
made is that the black-box function $f(\cdot)$ to be optimized has been generated from such a prior distribution, 
which is characterized by a zero mean and a covariance function $k(\mathbf{x},\mathbf{x}')$. 
That is $f(\cdot) \sim \mathcal{GP}(0,k(\cdot,\cdot))$.  The particular characteristics of $f(\cdot)$,
\emph{e.g.}, smoothness, additive noise, amplitude, etc., are specified by the covariance function 
$k(\mathbf{x},\mathbf{x}')$ which computes the covariance between $f(\mathbf{x})$ and $f(\mathbf{x}')$. 
A typical covariance function employed in the context of BO is the Mat\'ern 
function, in which the $\nu$ parameter is set equal to $3/2$ \citep{snoek2012}. This covariance function is:
\begin{align}
k(\mathbf{x},\mathbf{x}') & = \sigma^2 ( 1 + \frac{\sqrt{3}r}{\ell})\exp(-\frac{\sqrt{3}r}{\ell})\,,
\end{align}
where $r$ is the Euclidean distance between $\mathbf{x}$ and $\mathbf{x}'$. Namely, $|\mathbf{x}-\mathbf{x}'|$. 
Note that $k(\cdot,\cdot)$ only depends on $r$. This particular covariance function and others that share
this property are known as \textit{radial basis functions} (RBFs). $\ell$ is simply a hyper-parameter known as length-scale,
which controls the smoothness of the GP. Most of the times a different length scale $\ell_j$  is
used for each dimension $j$. $\sigma^2$ is the amplitude parameter, which controls the range of variability 
of the GP samples. Finally, $\nu$ is a hyper-parameter related to the number of times that the GP 
samples can be differentiated. Another popular covariance function is the squared exponential. 
In this case $k(\cdot,\cdot)$ is given by:
\begin{align}
k(\mathbf{x},\mathbf{x}') & = \sigma^2 \exp\left(-\frac{r^2}{2\ell^2}\right)\,.
\end{align}

Assume that we have already evaluated the objective at $N$ input locations. 
Let the corresponding data be summarized as $\mathcal{D}=\{(\mathbf{x}_i,y_i)\}_{i=1}^N$,
where $y_i = f(\mathbf{x}_i) + \epsilon_i$, with $\epsilon_i$ some additive Gaussian noise with variance $\sigma_0^2$.
A GP provides a predictive distribution for the potential values of $f(\cdot)$ at different regions
of the input space. This distribution is Gaussian 
and is characterized by a mean $\mu(\mathbf{x})$ and a variance 
$\sigma^2(\mathbf{x})$.  Namely, $p(f(\mathbf{x}^\star)|\mathbf{y}) =\mathcal{N}(f(\mathbf{x}^\star)|
m(\mathbf{x}^\star),  \sigma^2(\mathbf{x}^\star))$, where the mean and variance are respectively given by:
\begin{align}
\mu(\mathbf{x}) & = \mathbf{k}_{*}^{T} (\mathbf{K}+\sigma_{0}^{2}\mathbf{I})^{-1}\mathbf{y}\,, 
\label{eq:pred_mean}\\
\sigma^2(\mathbf{x}) & = k(\mathbf{x},\mathbf{x}) - \mathbf{k}_{*}^T(\mathbf{K}+\sigma_0^2\mathbf{I})^{-1}\mathbf{k}_*\,.
\label{eq:pred_var}
\end{align}
In the previous expression
$\mathbf{y}=(y_1,\ldots,y_{t-1})^\text{N}$ is a vector with the objective evaluations observed 
so far; $\mathbf{k}_*$ is a vector with the prior covariances between $f(\mathbf{x})$ and each $y_i$;
$\sigma_0^2$ is the variance of the additive Gaussian noise;
$\mathbf{K}$ is a matrix with the prior covariances among each $f(\mathbf{x}_i)$, for $i=1,\ldots,N$;
and $k(\mathbf{x},\mathbf{x})$ is the prior variance at the candidate location $\mathbf{x}$.
All these quantities are simply obtained by evaluating the covariance function $k(\cdot,\cdot)$ on the
corresponding input values. See \citep{rasmussen2005book} for further details.

BO methods use the previous predictive distribution to determine at which point $\mathbf{x}_{N+1}$ 
the objective function has to be evaluated. Once this new observation has been collected, the GP
model is updated with the new data and the process repeats. After collecting enough data like this,
the GP posterior mean given by (\ref{eq:pred_mean}) can be optimized to provide an estimate of the 
solution of the optimization problem. Notwithstanding, a GP has some hyper-parameters that need to be 
adjusted during the fitting process. These include the variance of the additive Gaussian noise 
$\sigma_0^2$, but also any potential hyper-parameter of the covariance function $k(\cdot,\cdot)$. 
These can be, \emph{e.g.}, the amplitude parameter and the length-scales \citep{rasmussen2005book}.  
Instead of finding point estimates for these hyper-parameters, an approach that has shown 
good empirical results is to compute an approximate posterior distribution for them using 
slice sampling \citep{snoek2012}. The previous Gaussian predictive distribution described in 
(\ref{eq:pred_mean}) and (\ref{eq:pred_var}) is then simply averaged over the hyper-parameter 
samples to obtain the final predictive distribution of the probabilistic model.

The key for BO success is found in the acquisition function $\alpha(\cdot)$. This function uses the predictive
distribution given by the GP to compute the expected utility of performing an evaluation of the objective
at each input location. The next point at which the objective has to be evaluated is simply $\mathbf{x}_{N+1} =
\text{arg max}_\mathbf{x} \,\, \alpha(\mathbf{x})$. Because this function only depends on the predictive distribution given by 
the GP and not on the actual objective $f(\cdot)$, the maximization of $\alpha(\cdot)$ is very cheap.
A popular acquisition function is expected improvement (EI) \citep{Jones98}. EI is given by 
the expected value of the utility function $u(y) = \text{max}{(0, \nu - y)}$ under the GP predictive distribution for $y$,
where $\nu=\text{min}(\{y_i\}_{i=1}^{N})$ is the best value observed so far, assuming minimization.
Therefore, EI measures on average how much we will improve on the current best found solution by performing 
an evaluation at each candidate point. The EI acquisition function is given by the following expression:
\begin{align}
\alpha(\mathbf{x}) = \sigma(\mathbf{x})(\gamma(\mathbf{x}) \Phi(\gamma(\mathbf{x}) + \phi(\gamma(\mathbf{x}))\,,
\end{align}
where $\gamma(\mathbf{x}) = (\nu - \mu(\mathbf{x})) /\sigma(\mathbf{x})$ and $\Phi(\cdot)$ and $\phi(\cdot)$ are
respectively the c.d.f. and p.d.f. of a standard Gaussian distribution.

Another popular acquisition function for BO is Predictive Entropy Search (PES) \citep{hernandez2014predictive}. PES is 
an information-theoretic method that chooses the next input location $\mathbf{x}_{N+1}$ at which the objective
function has to be evaluated as the one that maximizes the information about the global maximum $\mathbf{x}^\star$ of 
the optimization problem. The information about this minimum is given in terms of the differential 
entropy of the random variable $\mathbf{x}^\star$. This random variable is characterized by the corresponding 
posterior distribution $p(\mathbf{x}^\star|\mathcal{D}_N)$. PES simply chooses $\mathbf{x}_{N+1}$
as the point that maximizes the expected reduction in the differential entropy of $\mathbf{x}^\star$. 
The PES acquisition function is:
\begin{align}
\alpha(\mathbf{x}) = H[p(\mathbf{x}^\star|\mathcal{D}_N)] - \mathbb{E}_{y}[H[p(\mathbf{x}^\star|\mathcal{D}_N \cup (\mathbf{x},y))]\,,
\label{eq:pes}
\end{align}
where the expectation w.r.t. $y$ is given by the predictive distribution of the GP at $\mathbf{x}$.

A problem is, however, that evaluating (\ref{eq:pes}) in closed form is intractable. This expression has
to be approximated in practice. \cite{hernandez2014predictive} use the fact that the previous expression
is simply the mutual information between $\mathbf{x}^\star$ and $y$, $I(\mathbf{x}^\star;y)$, which is symmetric. 
Therefore one can swap the roles of $y$ and $\mathbf{x}^\star$ in (\ref{eq:pes}). This greatly simplifies the acquisition function
and an efficient approximation based on the expectation propagation algorithm is possible. PES has been
compared to other acquisition function showing improved optimization results. In particular, it has a better
trade-off between exploration and exploitation than EI.

\section{Dealing with Categorical and Integer-valued Variables} \label{sec:dealing}

In the framework described, the objective function $f(\cdot)$ 
is assumed to have input variables taking values on the real line. This is so, because in a GP the variables 
introduced in the covariance function $k(\cdot,\cdot)$ are assumed to be real. A problem 
may arise when some of the input variables can only take values in a closed subset of a discrete set, such as the 
integers, or when some the input variables are categorical. In this second case a typical approach
is to use a one-hot encoding of categorical variables. That is, the number of input dimensions is extended
by adding extra variables, one per potential category. The only valid configurations are those in which 
one of the extra variables takes value one (\emph{i.e.}, the extra variable corresponding to the active category), 
and all other extra variables take value zero.  For example, consider a categorical input dimension 
$x_j$ taking values in the set $\mathds{C}=\{\emph{red}, \emph{green}, \emph{blue}\}$.
We will replace dimension $j$ in $\mathbf{x}$ with three extra dimensional variables taking 
values $(1,0,0), (0, 1, 0)$ and $(0,0,1)$, for each value in $\mathds{C}$, respectively.

When not all input variables take real values, a standard GP will ignore that only some input variables configurations 
are valid and will place some probability mass on points at which $f(\cdot)$ cannot be evaluated. These incorrect 
modelling assumptions about $f(\cdot)$ may have a negative impact on the optimization process. 
Furthermore, the optimization of $\alpha(\cdot)$ will give candidate points $\mathbf{x}_{N+1}$
in which integer-valued or categorical variables will be assigned invalid values. In practice, some mechanism must 
be implemented to transform real values into integer or categorical values before the evaluation can take place. 
Importantly, if this is not done with care, some problems may appear.

\subsection{Naive and Basic Approaches}
\label{sec:basic_naive}

As described before, if the problem of interest considers some categorical or integer-valued variables, $f(\cdot)$
cannot be evaluated at all potential input locations. It can only be evaluated at those input locations that are 
compatible with the categorical or integer-valued variables. A naive approach to account for this 
is to (i) optimize $\alpha(\cdot)$ assuming all variables take values in the real line, and (ii) replace all 
the values for the integer-valued variables by the closest integer, and replace all categorical  variables 
with the corresponding one-hot encoding in which only one of the extra input variables takes value one and
all the others take value zero. In this second case, the active variable is simply chosen as the one 
with the highest value among the extra input variables. More precisely, let $\mathds{Q}_k$ be the set of 
extra input dimensions of $\mathbf{x}$ corresponding to the categorical input variable $k$ and let $j\in \mathds{Q}_k$. Then, we 
simply set $x_j = 1$ if $x_j > x_i$ $\forall i \in \mathds{Q}_k$ and $i \neq j$. Otherwise, $x_j = 0$. This is the 
approach followed by the popular software for BO Spearmint ({\small \url{https://github.com/HIPS/Spearmint}}). 

\begin{figure}[htb]
\begin{tabular}{l@{\hspace{1mm}}cc}
        \rotatebox{90}{\hspace{.7cm}{\bf \scriptsize Naive}} &
        \includegraphics[width=0.475\linewidth]{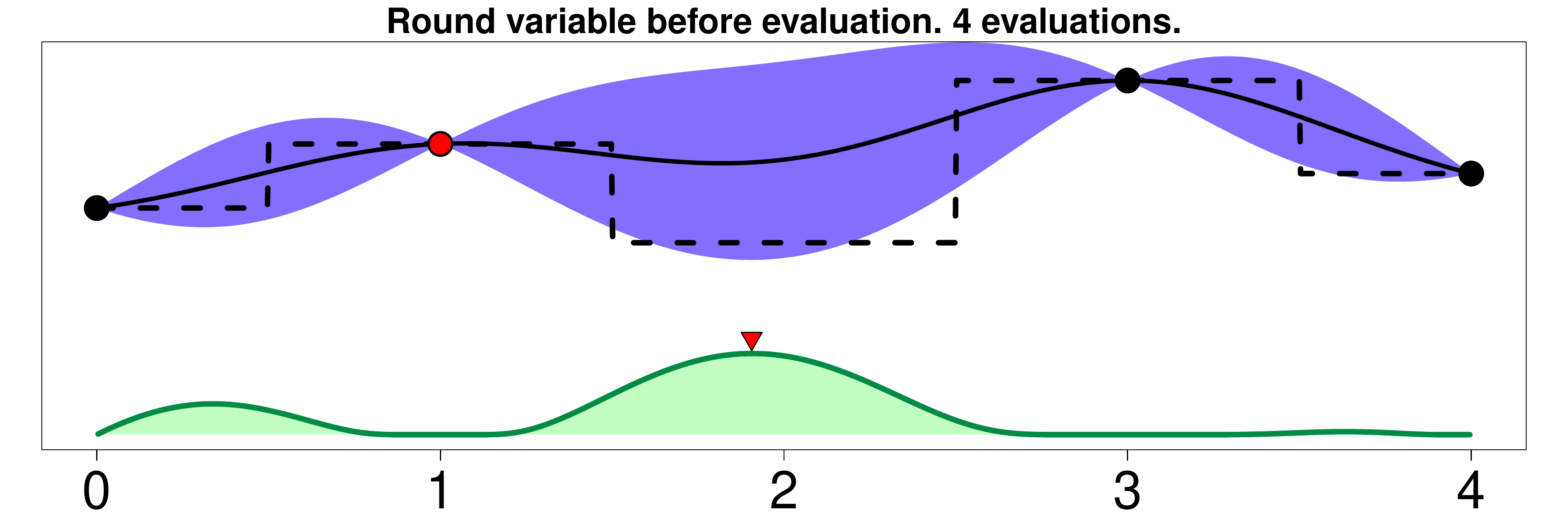} &
        \includegraphics[width=0.475\linewidth]{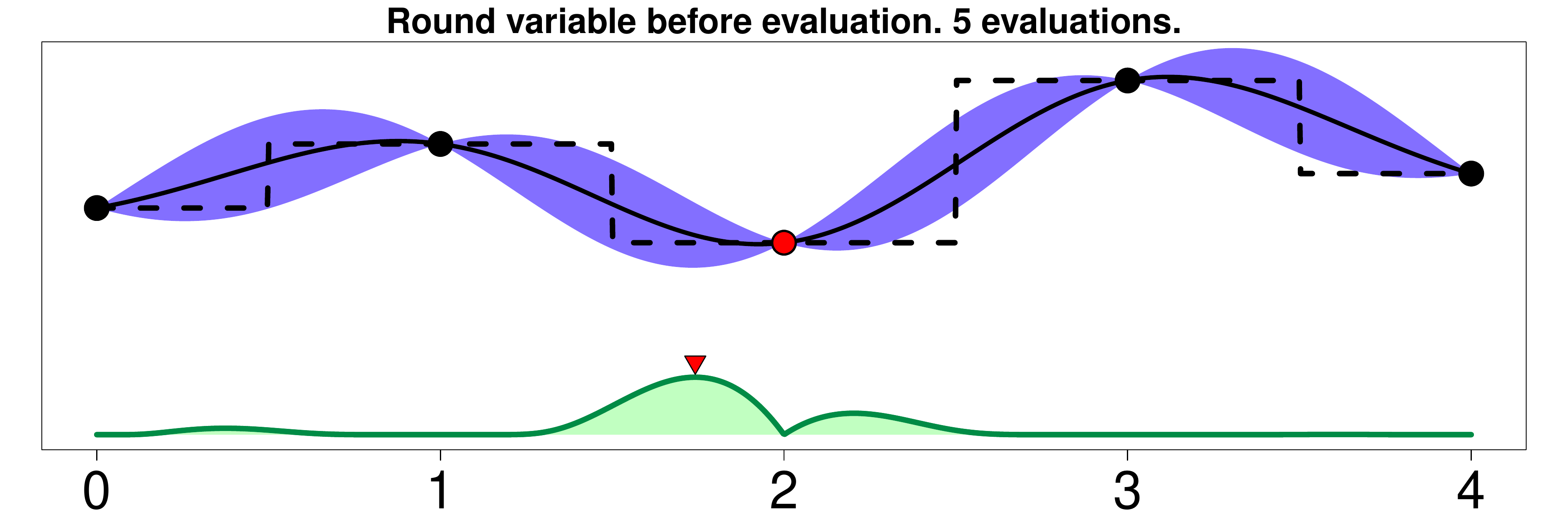} \\
        \rotatebox{90}{\hspace{.7cm}{\bf \scriptsize Basic}} &
        \includegraphics[width=0.475\linewidth]{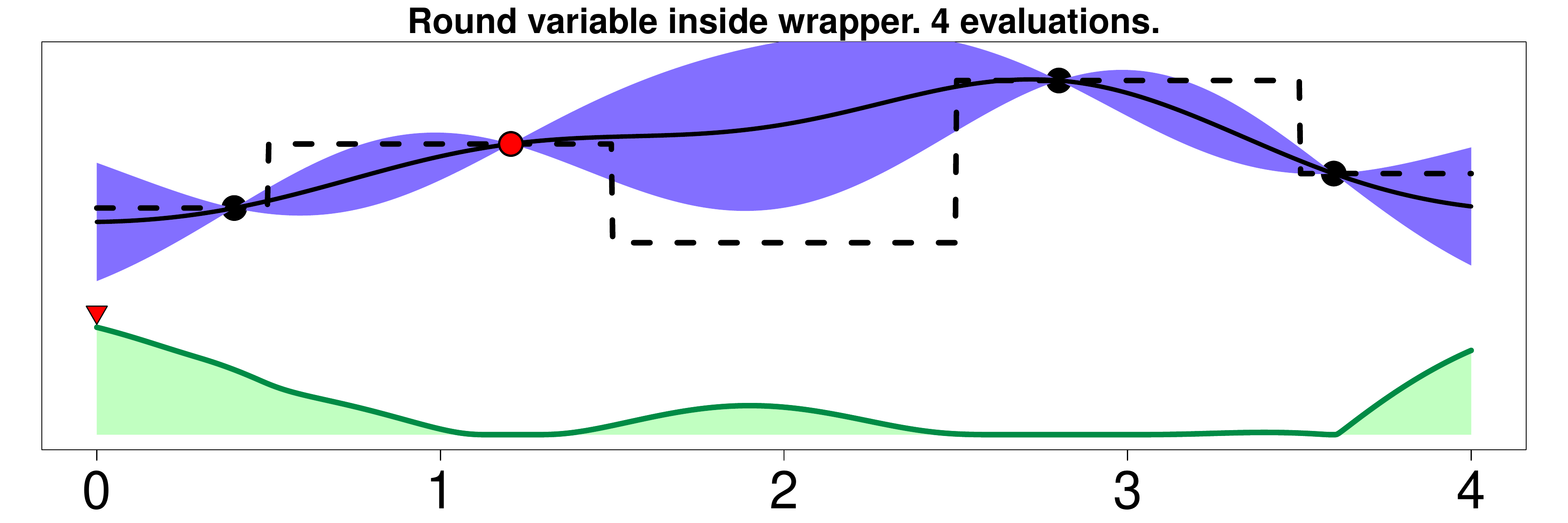} &
        \includegraphics[width=0.475\linewidth]{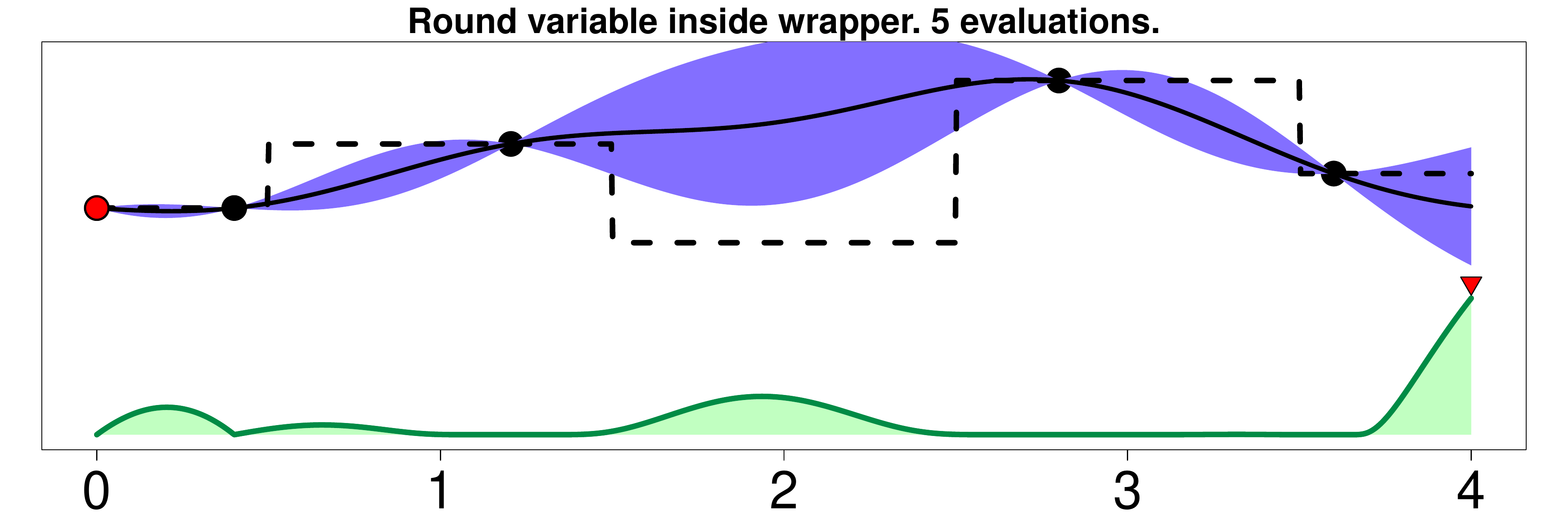} \\
        \rotatebox{90}{\hspace{.6cm}{\bf \scriptsize Proposed}} &
        \includegraphics[width=0.475\linewidth]{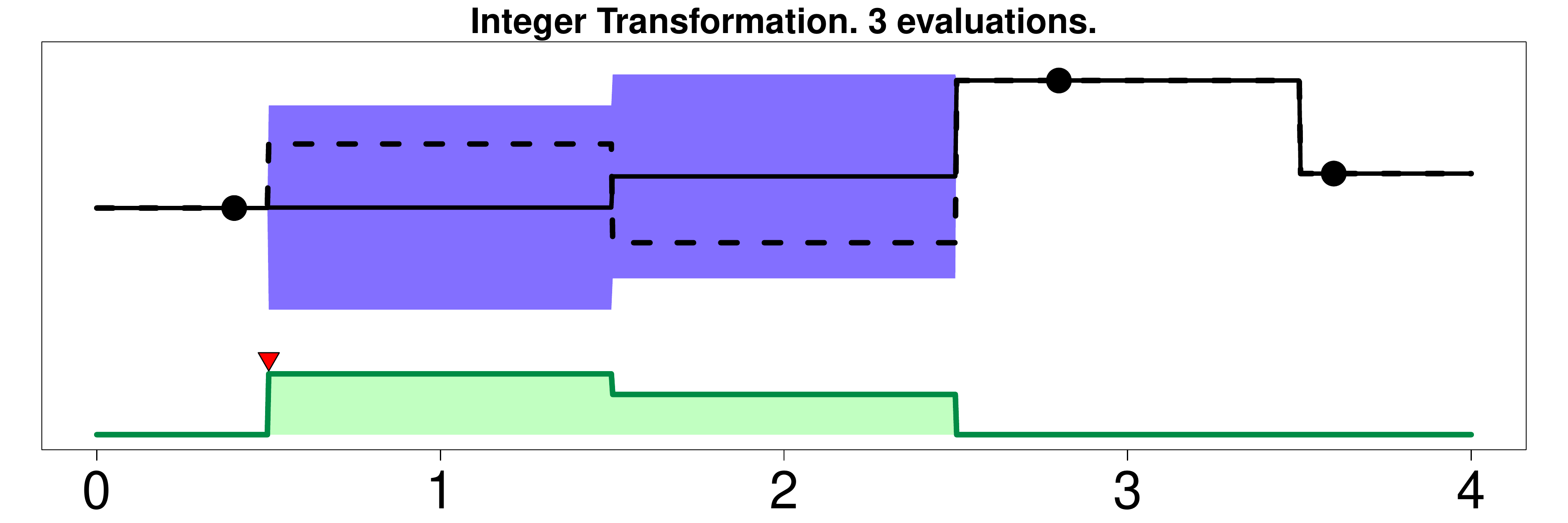} &
        \includegraphics[width=0.475\linewidth]{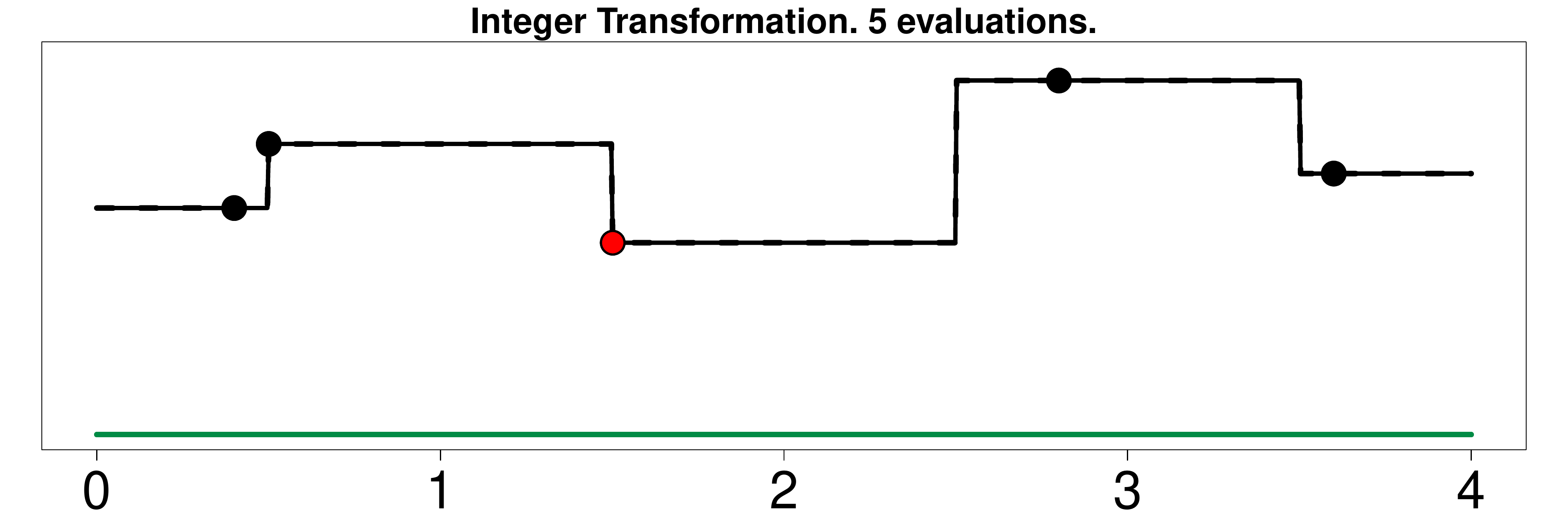} \\
\end{tabular}
\caption{{\small Different methods for dealing with integer-valued variables.
At the top of each image, we show a GP fit to the data (posterior mean and 1-std confidence interval, in purple)
that models a 1-dimensional objective taking values in the set $\{0,1,2,3,4\}$ (dashed line).
To display the objective we have rounded the real values at which to do the evaluation to the closest
integer. Below the GP fit, it is shown the acquisition function whose maximum is the recommendation for the next new
evaluation. Each column shows similar figures before and after evaluating a new point, respectively.
The proposed approach leads to no uncertainty about the objective after two evaluations.  Best seen in color.}}
\label{fig:methods}
\end{figure}

The first row of Figure \ref{fig:methods} shows, for an integer-valued input variable, that the naive approach 
just described can lead to a mismatch between the points in which the acquisition takes high values, and where the actual evaluation 
is performed. Importantly, this can produce situations in which the BO method always evaluates the objective 
at a point where it has already been evaluated. This may happen simply because the next and following 
evaluations are performed at different input locations from the one maximizing the acquisition function.
More precisely, since the evaluation is performed at a different point, it may not reduce at all the 
uncertainty about the potential values of the objective at the point maximizing the acquisition function.
Of course, in the case of categorical input variables this mismatch between the maximizer of the acquisition
function and the actual point at which the objective is evaluated will also be a problem. For this reason, 
we discourage the use of this approach.

The previous problem can be easily solved. In the case of integer-valued variables, one can simply do 
the rounding to the closest integer value inside the wrapper that evaluates the objective. 
In the case of categorical variables, a similar approach can be followed in inside the wrapper
using one-hot encoding. Namely, (i) look at which extra input variable has the largest
value, (ii) set that input variable equal to one, and (iii) set all other extra input variables equal to zero.
This basic approach is shown in the second row of Figure \ref{fig:methods} for the integer-valued case.
Here, the points at which the acquisition takes high values and the points at which the
objective is evaluated coincide. Thus, the BO method will tend to always evaluate at different input 
locations, as expected. This will avoid the problem described before, in which the BO method may get stuck.
The problem is, however, that the actual objective is constant in the intervals that are rounded to 
the same integer value. This constant behavior is ignored by the GP, and is expected to lead to sub-optimal
optimization results. The same behavior is expected in the case of categorical input variables.

\subsection{Proposed Approach} \label{sec:integer}

We propose a method to alleviate the problems of the basic approach described in Section \ref{sec:dealing}. 
For this, we consider that the objective should be constant in those regions of the input space that
lead to the same input variable configuration on which the actual objective has to be evaluated. 
This property can be easily introduced into the GP by modifying
$k(\cdot,\cdot)$. Covariance functions are often stationary and only depend on the distance between the 
input points \citep{rasmussen2005book}. If the distance between two points is zero, the values of the function at 
both points will be the same (the correlation is equal to one). Based on this fact, we suggest to transform 
the input points to $k(\cdot,\cdot)$, obtaining an alternative covariance function $k'(\cdot,\cdot)$:
\begin{align}
k'(\mathbf{x}_i,\mathbf{x}_j) &= k(T(\mathbf{x}_i),T(\mathbf{x}_j)) \,,
\label{eq:covariance}
\end{align}
where $T(\mathbf{x})$ is a transformation in which all non real input variables of $f(\cdot)$ in $\mathbf{x}$ 
are modified as follows:
\begin{itemize}
\item The input variables corresponding to an integer-valued input variable are rounded to the closest integer value.
\item All extra input variables corresponding to the same categorical input variable are assigned zero value unless
	they take the largest value among the corresponding group of extra variables. If they take the largest value,
	they are assigned value one.
\end{itemize}
Essentially $T(\cdot)$ does the same transformation on $\mathbf{x}$ as the one described in Section
\ref{sec:basic_naive} for the basic approach inside the wrapper that evaluates the objective. Importantly, however, 
this transformation also takes place in the covariance function of the GPs, which will allow for a 
better modeling of the objective. 

The beneficial properties of $k'(\cdot,\cdot)$ when used for BO are illustrated in the 
third row of Figure \ref{fig:methods} for the case of an integer-valued input variable. 
We can see that the GP model correctly identifies that the objective function is constant 
inside intervals of real values that are rounded to the same integer. The uncertainty is also 
the same in those intervals, and this is reflected in the acquisition function. Furthermore, after performing 
a single measurement in each interval, the uncertainty about $f(\cdot)$ goes to zero. This better modelling
of the objective is expected to be reflected in a better performance of the optimization process.
The same behavior is expected in the case of categorical variables.

In the case of integer-valued variables, the transformation $T(\mathbf{x})$ will round all integer-valued 
variables values in $\mathds{R}$ to the closest integer $k \in \mathds{Z}$. The set of integer values, 
$\mathds{Z}$, has a notion of order. That is, for all $z \in \mathbb{Z}$, we can define operators of order 
that involve two values: $<,>,\leq$ and $\geq$, such that $z_i < z_j$, $z_j > z_i$, $z_i \leq z_j$ and 
$z_j \geq z_i$, having that $z_i,z_j \in \mathbb{Z}$. This order will be preserved by the
resulting transformation. More precisely, assume an integer input variable and that $T(\mathbf{x})$ and $T(\mathbf{x}')$ 
only differ in the value of such integer input variable. The prior covariance between 
$f(\mathbf{x})$ and $f(\mathbf{x}')$ under $k(T(\mathbf{x}),T(\mathbf{x}'))$ will be higher the closer the 
corresponding integer values of $T(\mathbf{x})$ and $T(\mathbf{x}')$ are one from another. Therefore, the GP 
will be able to exploit the smoothness in the objective $f(\cdot)$ when solving the optimization problem.

In the case of categorical variables (\emph{e.g.}, variables that can take values such as 
\emph{red}, \emph{green}, \emph{blue}) there is no notion of order. That is, the operators 
$<,>,\geq$ and $\leq$ have no meaning nor purpose. One can not compare two different 
values $c_1,c_2$ of any categorical-valued set $\mathds{C}$ according to these operators. However, 
what does exist in a categorical set is a notion of equality or difference, given by the operators $=,\neq$. 
The proposed transformation is able to preserve this notion of no order and notion of equal or 
different. More precisely, assume a single categorical variable and that $T(\mathbf{x})$ and $T(\mathbf{x}')$ 
only differ in the values of the corresponding extra variables associated to that categorical variable.
The prior covariance between $f(\mathbf{x})$ and $f(\mathbf{x}')$ under 
$k(T(\mathbf{x}),T(\mathbf{x}'))$ will be the same as long as $T(\mathbf{x})$ and $T(\mathbf{x}')$
encode a different value for the categorical variable. Of course if $T(\mathbf{x})$ and $T(\mathbf{x}')$
encode the same value for the categorical variable, the covariance will be maximum.

\begin{figure}[htb]
\begin{center}
\begin{tabular}{lcc}
	\rotatebox{90}{\hspace{1.75cm}{\bf \scriptsize Proposed Approach}} &  
        \includegraphics[width=0.375\linewidth]{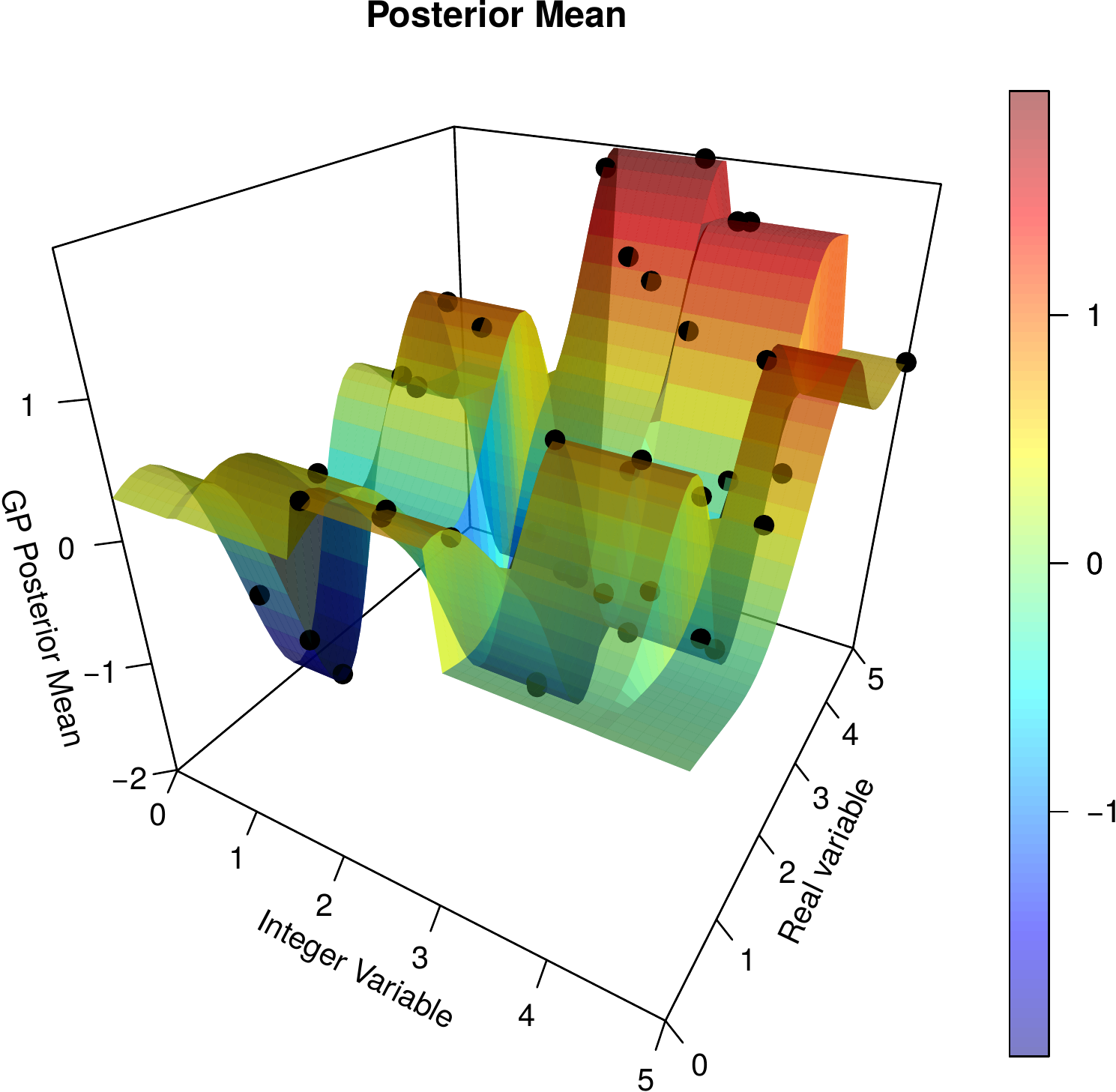} &
        \includegraphics[width=0.375\linewidth]{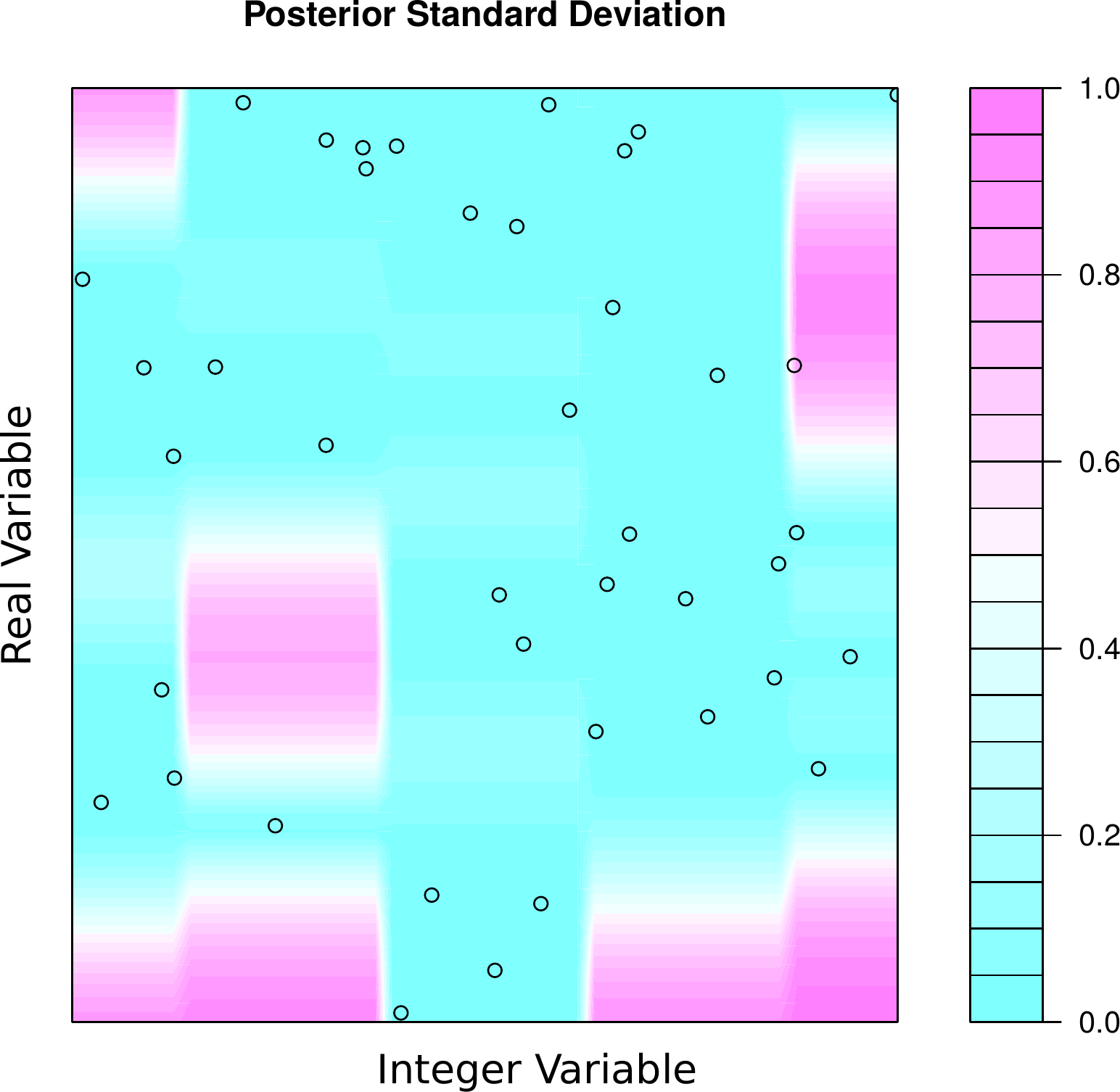} \\
	\rotatebox{90}{\hspace{2.25cm}{\bf \scriptsize Standard GP}} &  
        \includegraphics[width=0.375\linewidth]{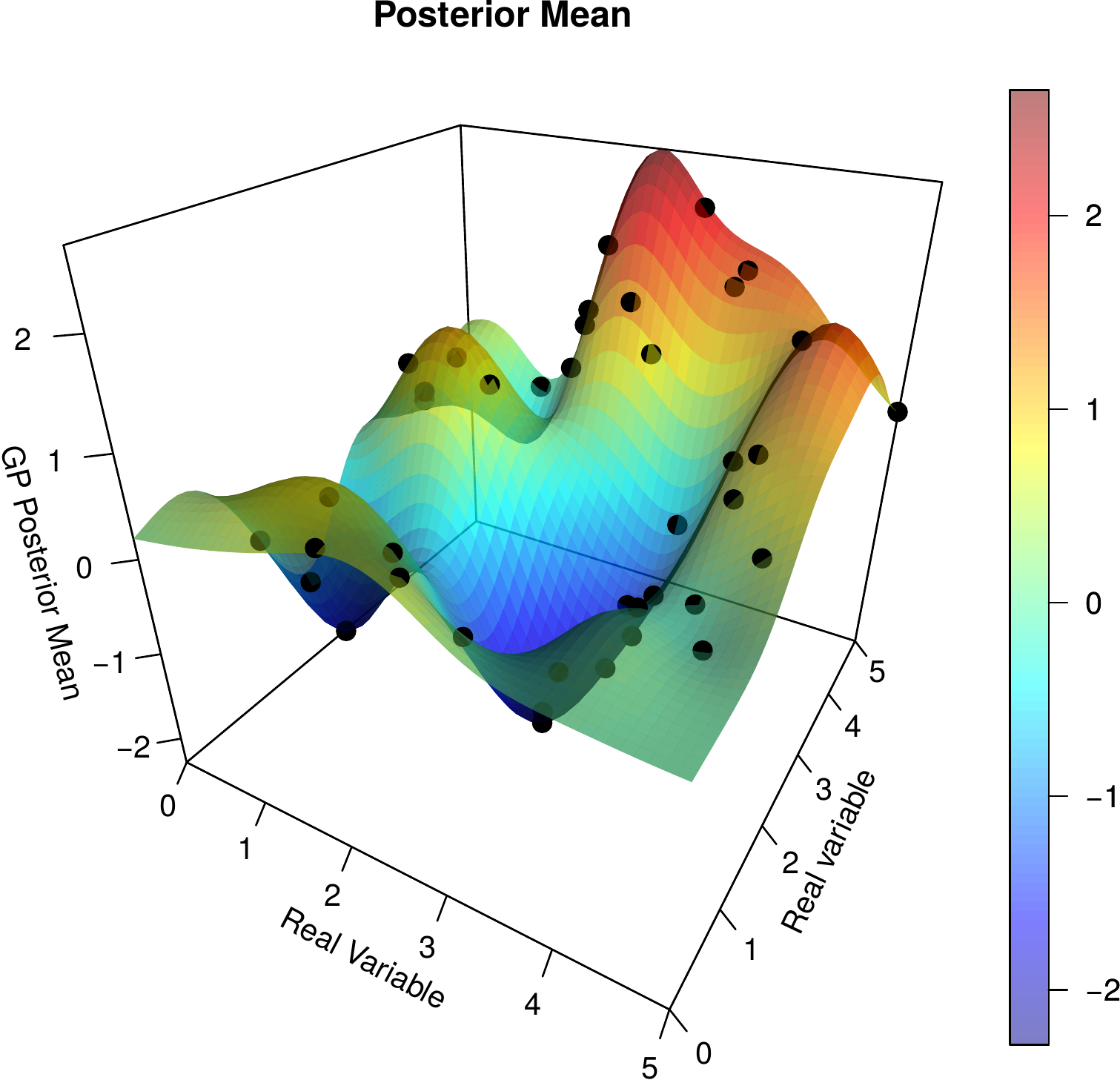} &
        \includegraphics[width=0.375\linewidth]{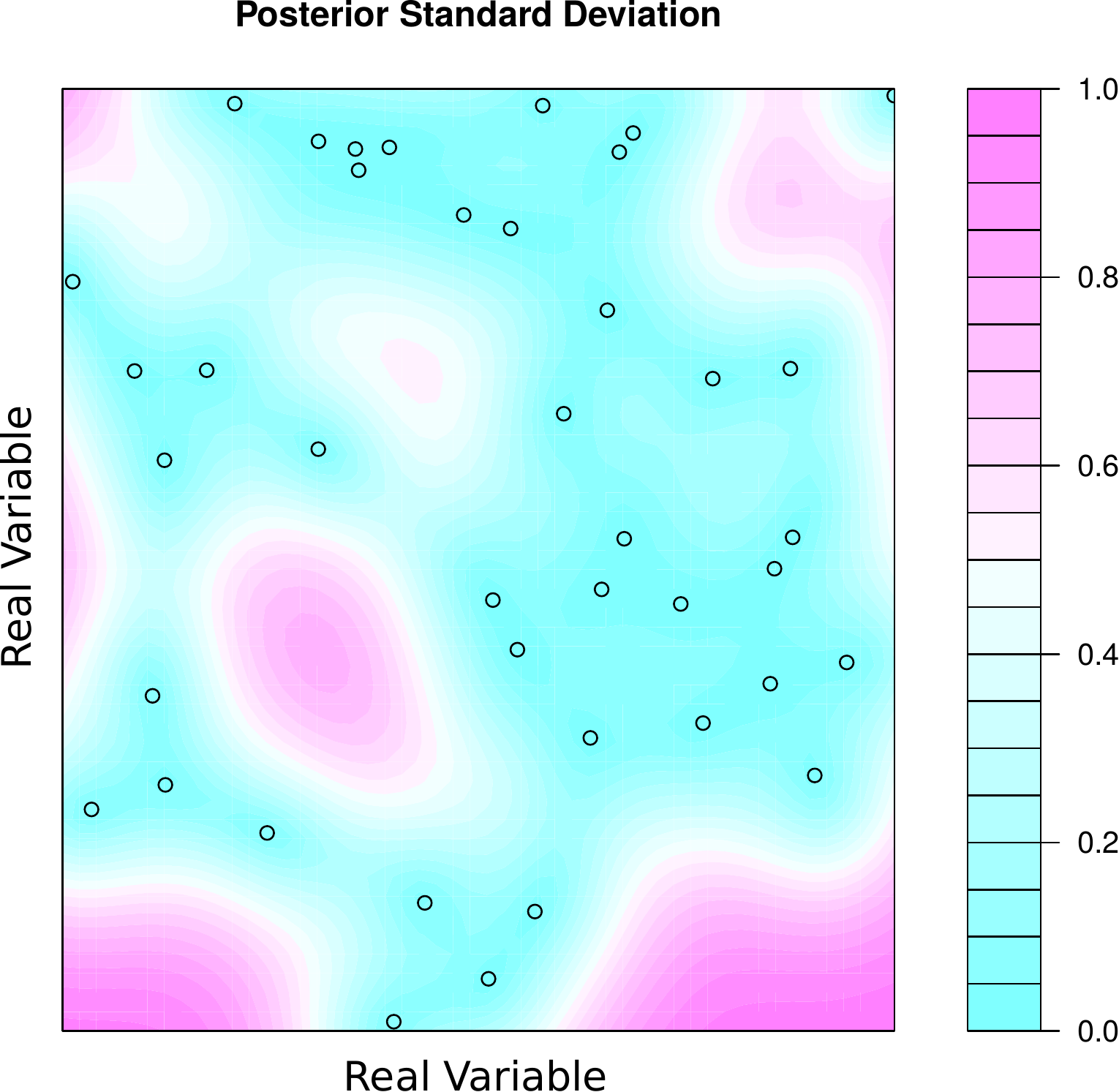} \\
\end{tabular}
\end{center}
\caption{{\small (top) Posterior mean and standard deviation of a GP model over a 2-dimensional space in which the first dimension
can only take $5$ different integer values and when the covariance function in (\ref{eq:covariance}) is used. Note that the second 
dimension can take any real value. (bottom) Same results for a GP model using a covariance function without the proposed
transformation.  Best seen in color.}}
\label{fig:posterior}
\end{figure}

Figure \ref{fig:posterior} illustrates the described modelling properties in the case of a real 
and an integer-valued variable in the transformation suggested in (\ref{eq:covariance}). 
This figure shows the mean and standard deviation of the posterior 
distribution of a GP given some observations. The results obtained with a standard GP that does not use the proposed 
transformation are also displayed. In this case, the data has been sampled from a GP using the covariance function in (\ref{eq:covariance}) 
where $k(\cdot,\cdot)$ is the squared exponential covariance function \citep{rasmussen2005book}. One dimension takes continuous values 
and the other dimension takes values in $\{0,1,2,3,4\}$. Note that the posterior distribution captures the constant 
behavior of the function in any interval of values that are rounded to the same integer, only for the integer 
dimension (top). A standard GP (corresponding to the basic approach in Section \ref{sec:dealing}) cannot capture 
this shape (bottom).

\begin{figure}[htb]
\begin{center}
\begin{tabular}{lcc}
        \rotatebox{90}{\hspace{1.75cm}{\bf \scriptsize Proposed Approach}} &
        \includegraphics[width=0.375\linewidth]{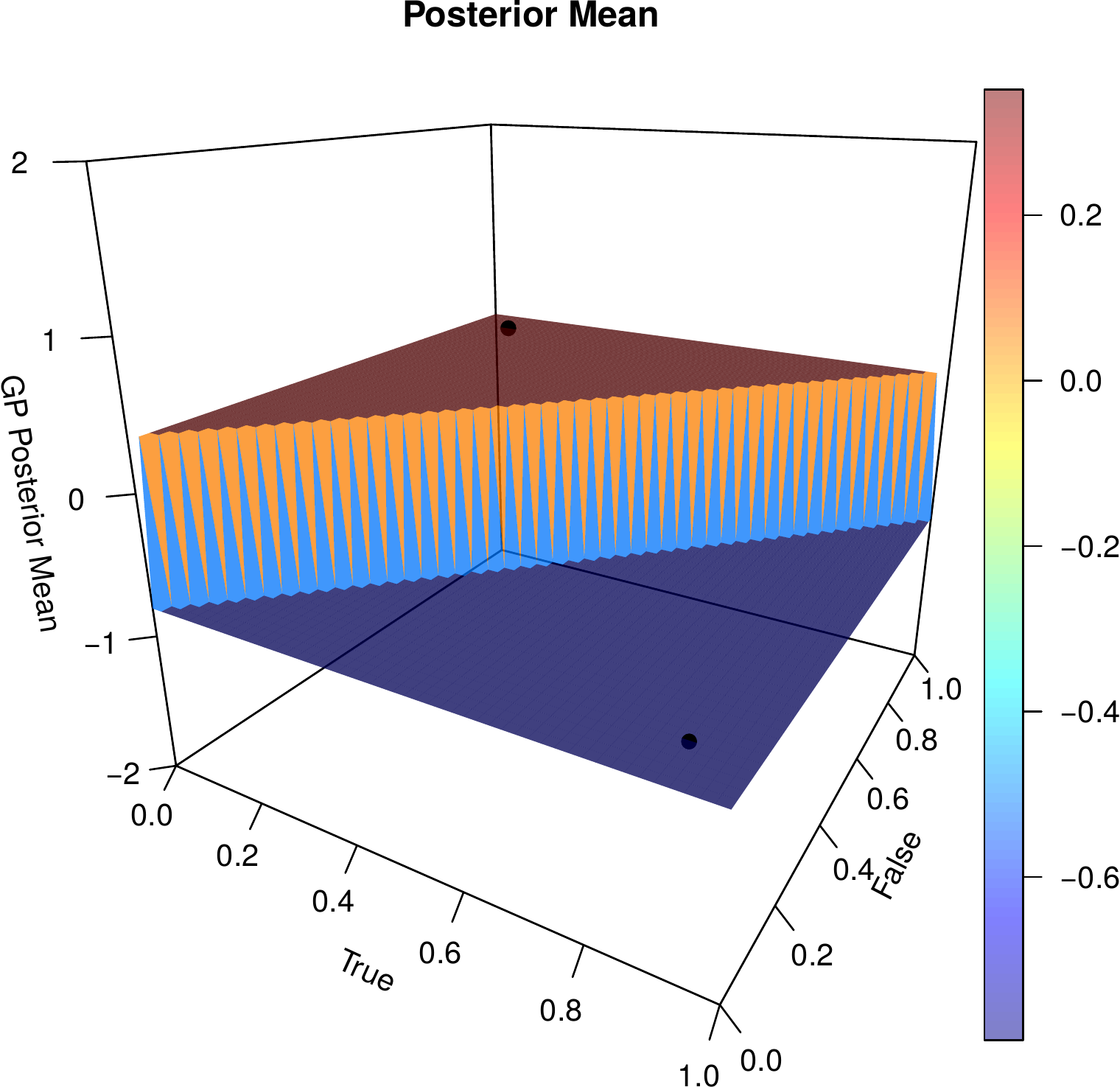} &
        \includegraphics[width=0.375\linewidth]{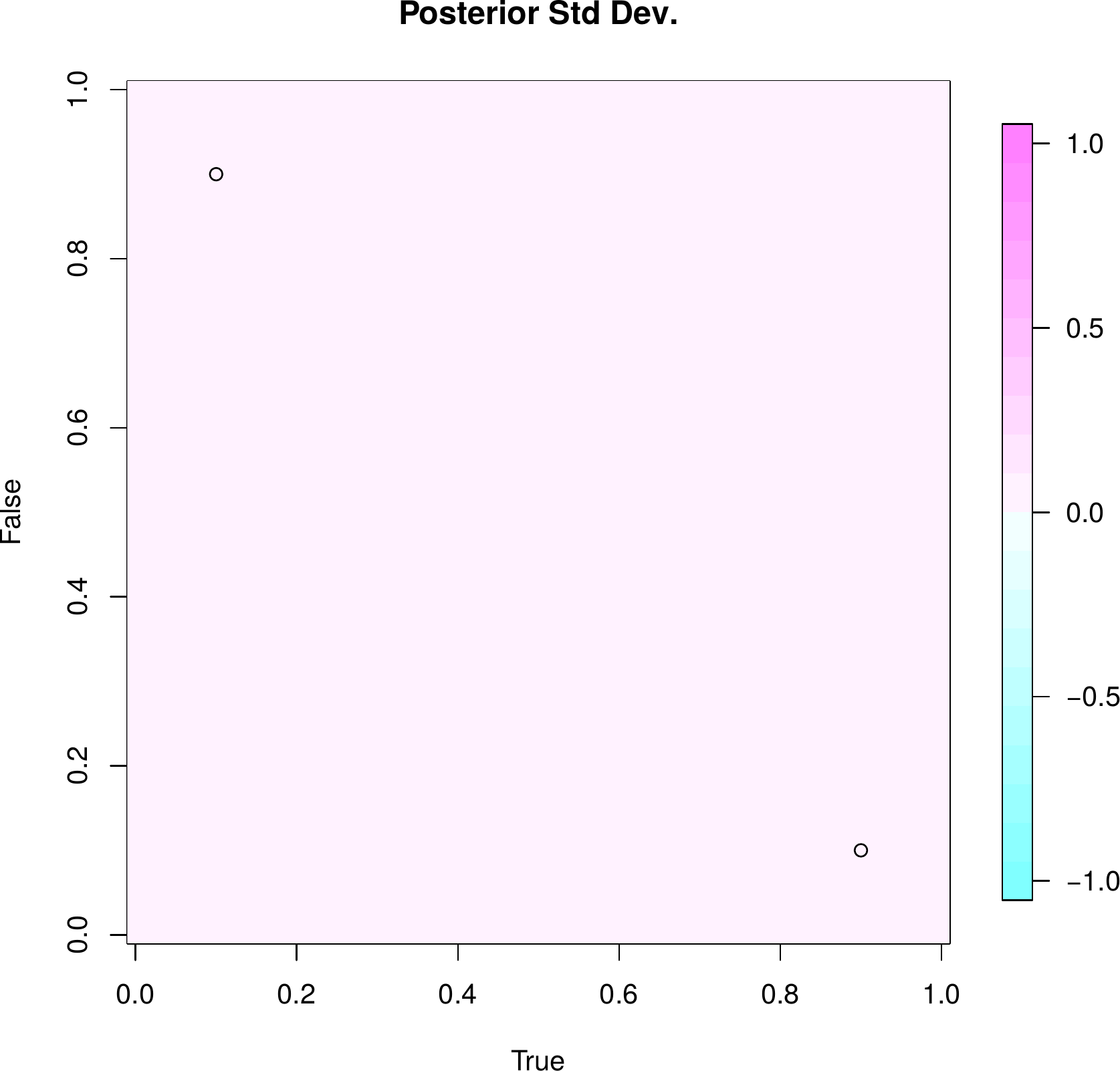} \\
        \rotatebox{90}{\hspace{2.25cm}{\bf \scriptsize Standard GP}} &
        \includegraphics[width=0.375\linewidth]{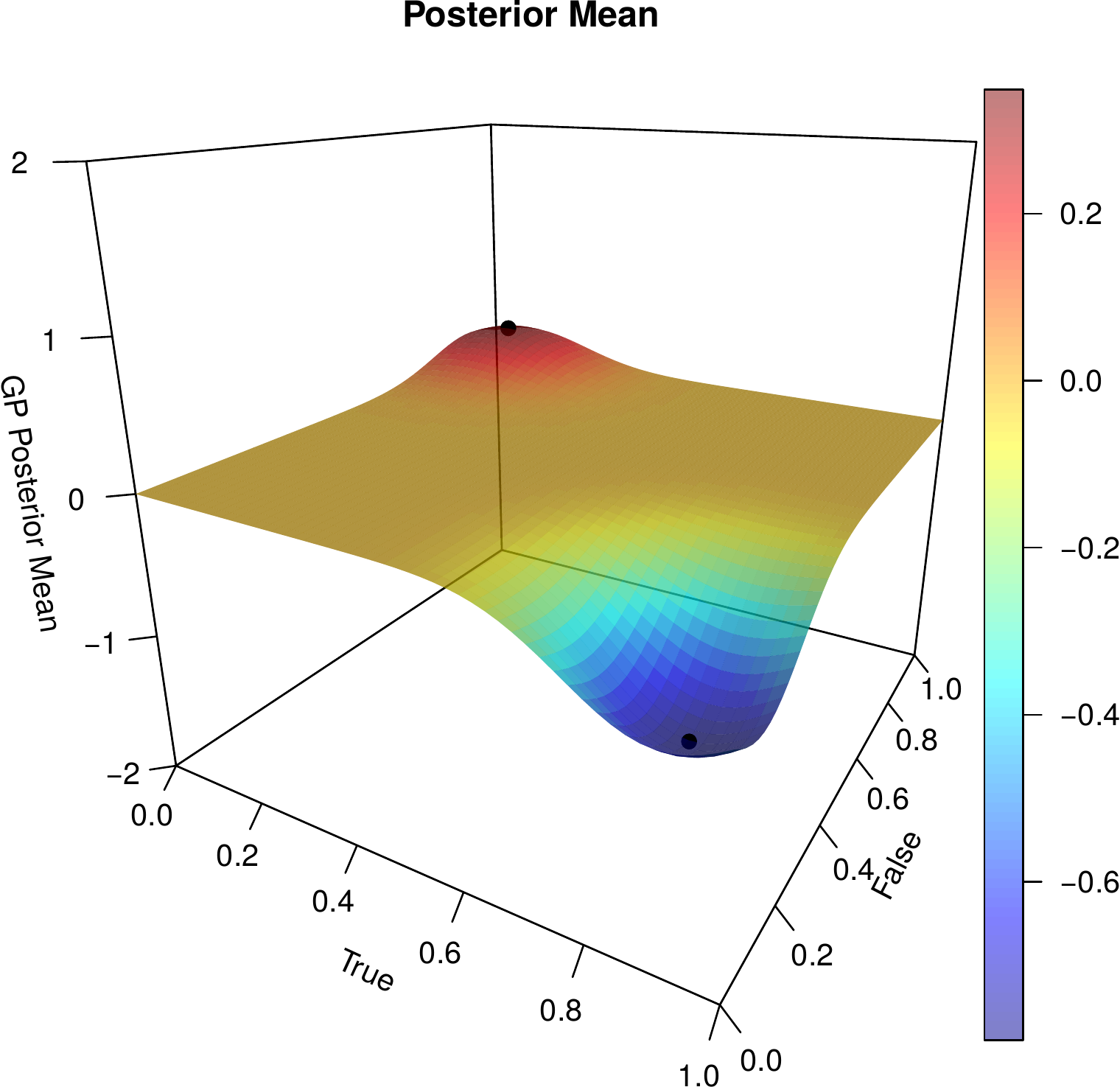} &
        \includegraphics[width=0.375\linewidth]{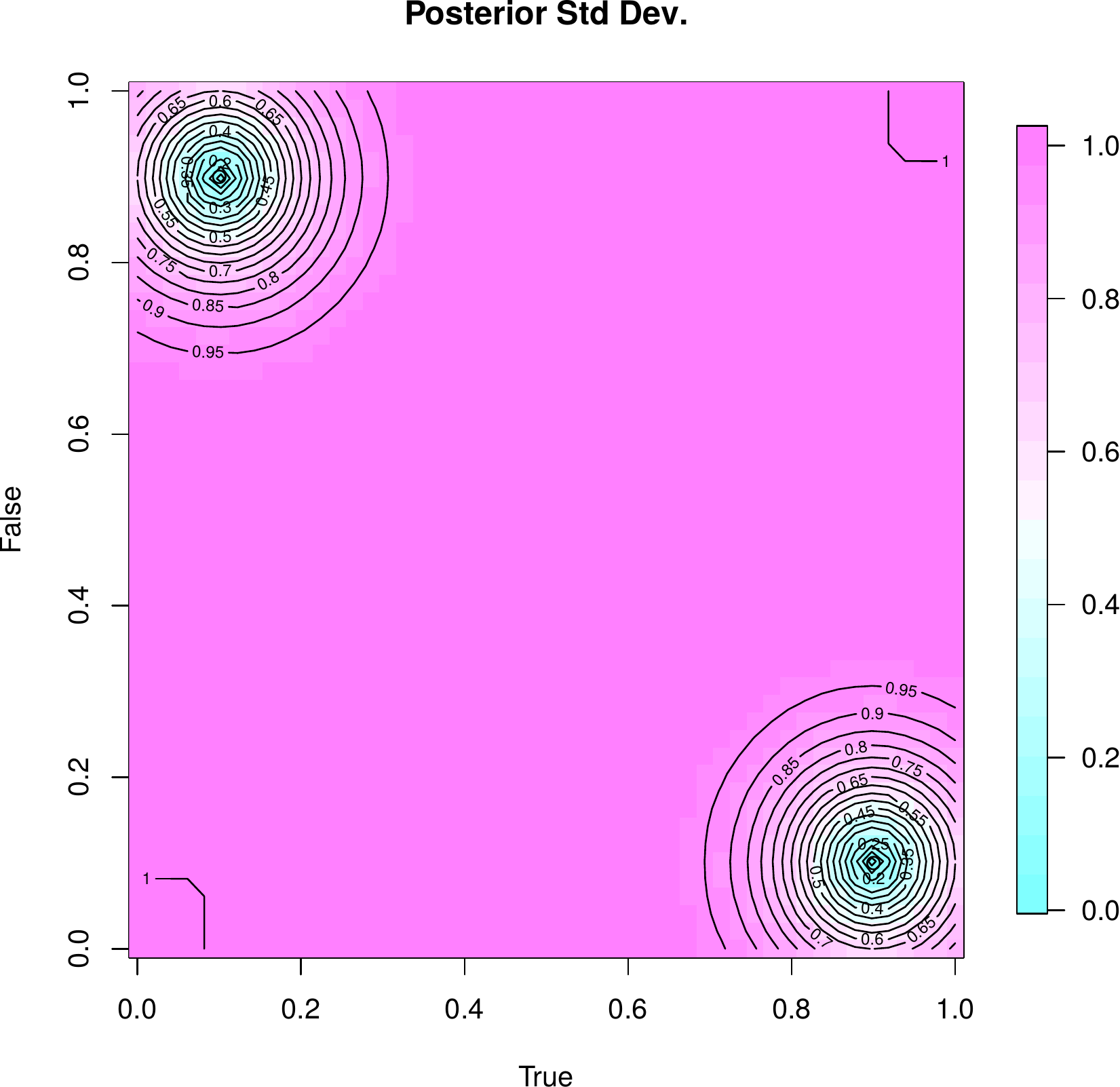} \\
\end{tabular}
\end{center}
\caption{{\small (top) Posterior mean and standard deviation of a GP model over a 1-dimensional categorical variable 
taking only two different values. The covariance function in (\ref{eq:covariance}) is used in the GP model in this case. 
Note that in the uncertainty goes to zero after these two single observations. Same results using a covariance function 
without the proposed transformation. In this case the uncertainty about the potential values of the function is very big.
 Best seen in color.}}
\label{fig:posterior_categorical}
\end{figure}

Figure \ref{fig:posterior_categorical} illustrates the proposed transformation for the categorical case and a 
single variable that can only take two values, \emph{e.g.}, \emph{True} and \emph{False}. Using one-hot encoding,
these two values will be represented as $(0,1)$ and $(1,0)$, respectively. In the naive approach described before, this 
categorical variable will be replaced by two real variables taking values in the range $[0,1]$. Notwithstanding, any 
combination of values in which the first component is larger than the second will lead to the configuration value $(1,0)$.
Conversely, any combination of values in which the second component is larger will lead to the configuration value $(0,1)$.
Therefore, the corresponding objective will be constant in those regions of the input space that lead to the same configuration. 
This behavior is illustrated by Figure \ref{fig:posterior_categorical} top, in which the posterior distribution of the GP is plotted 
given two observations. In this case we use the proposed transformation of the covariance function.
Note that the uncertainty goes to zero after just having a single observation corresponding to 
the \emph{True} value and a single observation corresponding to the \emph{False} value. This makes sense, because 
the objective is constant in all those regions of the input space that lead to the same configuration of 
the extra variables introduced in the input space. In Figure \ref{fig:posterior_categorical} (bottom) we show 
that a standard GP cannot model this behavior, and the posterior distribution of the mean is not constant in 
those regions of the input space that lead to the same configuration for the categorical variable. 
Furthermore, the posterior standard deviation is significantly different from zero, unlike in the proposed approach. 
Summing up, Figure \ref{fig:posterior_categorical} shows that by using the proposed covariance function, we are
better modeling the objective function, which in the end will be translated in better optimization results.

\section{Related Work} \label{sec:related}

We describe here two approaches that can be used as an 
alternative to BO methods using GPs when categorical and/or
integer-valued variables are present in a black-box optimization problem. These 
are Sequential model-based optimization for general algorithm configuration (SMAC) 
\citep{hutter2011sequential} and the Tree-structured Parzen Estimator Approach (TPE) 
\citep{bergstra2011algorithms}. Both can naturally handle integer and categorical-valued 
variables. SMAC is present in the popular machine learning tool AutoWeka \citep{thornton2013auto}. 
TPE is used in the HyperOpt tool \citep{bergstra2013hyperopt}.  

SMAC uses a random forest as the underlying surrogate model of the black-box 
objective \citep{breiman2001random}. The predictive distribution given by 
this model is used to select promising parameter values on which the objective should 
be evaluated. In random forest $T$ random regression trees are iteratively fit using 
each time a bootstrap sample of training data. Each bootstrap sample is obtained by drawing 
with replacement from the observed data $N$ instances. Furthermore, in random forest, at each node, 
a randomly chosen subset of variables are tested to split the data. This introduces variability in 
the generated regression trees. Given a candidate test location, the prediction for 
that point is computed for each of the $T$ trees. The predictive distribution of the model
is simply a Gaussian distribution with the empirical mean and variance across the individual 
tree predictions. Given this predictive distribution, the EI criterion described in Section 
\ref{sec:back} is computed and used to select a new point at which the objective $f(\cdot)$ should 
be evaluated. The main advantage of random forest is that it has a smaller computational cost than 
a GP.

The regression trees used by random forest to compute the predictive 
distribution can naturally consider integer and categorical-valued variables.
Therefore this method does not suffer from the limitations described in Section \ref{sec:dealing}
for GPs. A problem, however, is that the predictive distribution of random forest is not 
very good. In particular, it relies on the randomness introduced by the bootstrap samples 
and the randomly chosen subset of variables to be tested at each node to split the data. 
This result is confirmed by our experiments, in which BO methods using GPs tend to 
perform better than SMAC.

TPE also uses EI as the acquisition function. However, its computation is carried out in a 
different way, using a different modelling strategy. Whereas standard BO methods fit a 
discriminative model for $p(y|\mathbf{x})$ directly, TPE follows a generative approach. More 
precisely, $p(\mathbf{x}|y)$ and $p(y)$ are fit instead. Both approaches are related as 
$p(y|\mathbf{x}) = \frac{p(\mathbf{x}|y)p(y)}{p(\mathbf{x})}$ where $p(\mathbf{x}) = \int p(\mathbf{x}|y)p(y)dy$. 
To obtain an estimate of  $p(\mathbf{x}|y)$, TPE models each dimension with a probability distribution that serves 
as a prior for that dimension. Then, TPE replaces that distributions with non-parametric densities. 
TPE redefines $p(\mathbf{x}|y)$ by using two different densities, $\ell(\mathbf{x})$ and $g(\mathbf{x})$.
$\ell(\mathbf{x})$ is estimated using 
the observations in which the evaluation is lower than a chosen value $y^\star$. $g(\mathbf{x})$ is estimated 
using the rest of observations, respectively. That is,
\begin{align}
p(\mathbf{x}|y) & = \left\{ \begin{array}{cc}
	\ell(\mathbf{x}) & \text{if} \quad y \leq y^\star \,, \\
	g(\mathbf{x}) & \text{if} \quad y > y^\star \,.  
\end{array} \right.
\end{align}
Importantly, these two densities are obtained using Parzen estimators, a non-parametric density estimator, 
in the case of continuous random variables. In the case of  categorical
variables, a categorical distribution is used instead. Similarly, in the case of a variable over the integers,
a distribution that considers only this domain is used instead. This can easily account for categorical and integer-valued 
input variables in TPE. $y^\star$ is simply set as some quantile of the observed $y$ values. 
An interesting property of this approach is that no specific model for $p(y)$ is necessary. 
TPE derives a different expression for the EI acquisition function. Namely,
\begin{align}
\alpha(\mathbf{x}) = \int_{-\infty}^{y^\star}(y^\star-y)p(y|\mathbf{x})dy = \int_{-\infty}^{y^\star}
	(y^\star-y)\frac{p(\mathbf{x}|y)p(y)}{p(\mathbf{x})} dy \propto (\gamma + \frac{g(\mathbf{x})}{\ell(\mathbf{x})}(1-\gamma))^{-1}\,,
\end{align}
where we have used that $\gamma = p(y<y^\star)$ and that $p(\mathbf{x}) = \int p(\mathbf{x}|y)p(y)dy = \gamma l(\mathbf{x}) +
(1-\gamma)g(\mathbf{x})$. See \citep{bergstra2011algorithms} for further detail.
The TPE EI criterion is hence maximized simply by choosing points with high probability 
under $\ell(\mathbf{x})$ and low probability under $g(\mathbf{x})$. 
The advantage of this approach is that both of the models, $\ell(\mathbf{x})$ and $g(\mathbf{x})$, are 
hierarchical processes that naturally take into account discrete-valued and continuous-valued variables.

In the literature there are other approaches for BO with GPs that can account for categorical and 
integer-valued input variables \citep{rainforth2016bayesian}. However, they follow 
the basic approach described in Section \ref{sec:dealing} and are expected to be sub-optimal when compared 
to the proposed approach. Therefore, we do not consider them in our experiments.

\section{Experiments} \label{sec:experiments}

We carry out several experiments to evaluate the performance of the proposed approach for considering
integer and categorical-valued input variables in BO methods using GPs. We compare its performance with 
that of the basic approach described in Section \ref{sec:dealing}. Each of these two approaches has been 
implemented in the software for BO Spearmint (\url{https://github.com/HIPS/Spearmint}.  We also 
compare results with other methods that do not use a GP as a surrogate model,
in both synthetic and real problems. Those methods are
the ones described in Section \ref{sec:related}. Namely, SMAC and TPE.

The results reported in this section show averages and corresponding standard deviations obtained
after 100 repetitions of each experiment. Standard deviations are estimated using 
200 bootstrap samples. In the synthetic problems we consider two scenarios. Namely, noiseless and noisy
observations, and report results for both of them. The hyper-parameters of each GP
in the basic and the proposed approaches (length-scales, level of noise and amplitude) are 
approximately sampled from their posterior distribution using slice sampling as in \citep{snoek2012}. 
In the synthetic experiments, we generate 10 samples for each hyper-parameter, and the acquisition function of 
each method is averaged over these samples. In the real-world problems we generate $50$ samples for each hyper-parameter. 
We use a Mat\'ern covariance function in the GPs.
In the synthetic experiments,
at each iteration of the optimization process, each method outputs a recommendation 
obtained by optimizing the mean prediction of the GP. In the real world-experiments,
however, we simply output the best-observed evaluation, to reduce the impact of model bias. 
SMAC and TPE approaches deliver a recommendation based on the best-observed 
evaluation, in both synthetic and real-world scenarios. 

\subsection{Synthetic Experiments}

A first batch of experiments considers minimizing synthetic objectives which are sampled from a GP prior.
More precisely, we generate 100 optimization problems in $2$ and $4$ dimensions.
We consider two scenarios. Each scenario involves optimizing a combination of real and categorical input 
variables, and a combination of real and integer-valued input variables, respectively. 
These two scenarios are also analyzed in the case of noiseless and noisy evaluations.
The variance of the additive Gaussian noise is set equal to $0.01$.
Here we use EI as the acquisition function for the basic and the proposed approaches.
We run each method (Basic, Proposed, SMAC and TPE) for 50 iterations in these experiments.

Note that SMAC and TPE do not assume a GP for the underlying model and could be in disadvantage in these 
experiments. However, we believe it is still interesting to compare results with them in this setting in which 
the exact solution of the optimization problem can be easily obtained and the level of noise can be controlled. 
In the following section we carry out experiments in which the actual objectives need not be sampled from a GP,
to illustrate the advantages of the proposed approach in a wider range of problems.

\begin{figure}[htb]
\begin{tabular}{cc}
        \includegraphics[width=0.475\linewidth]{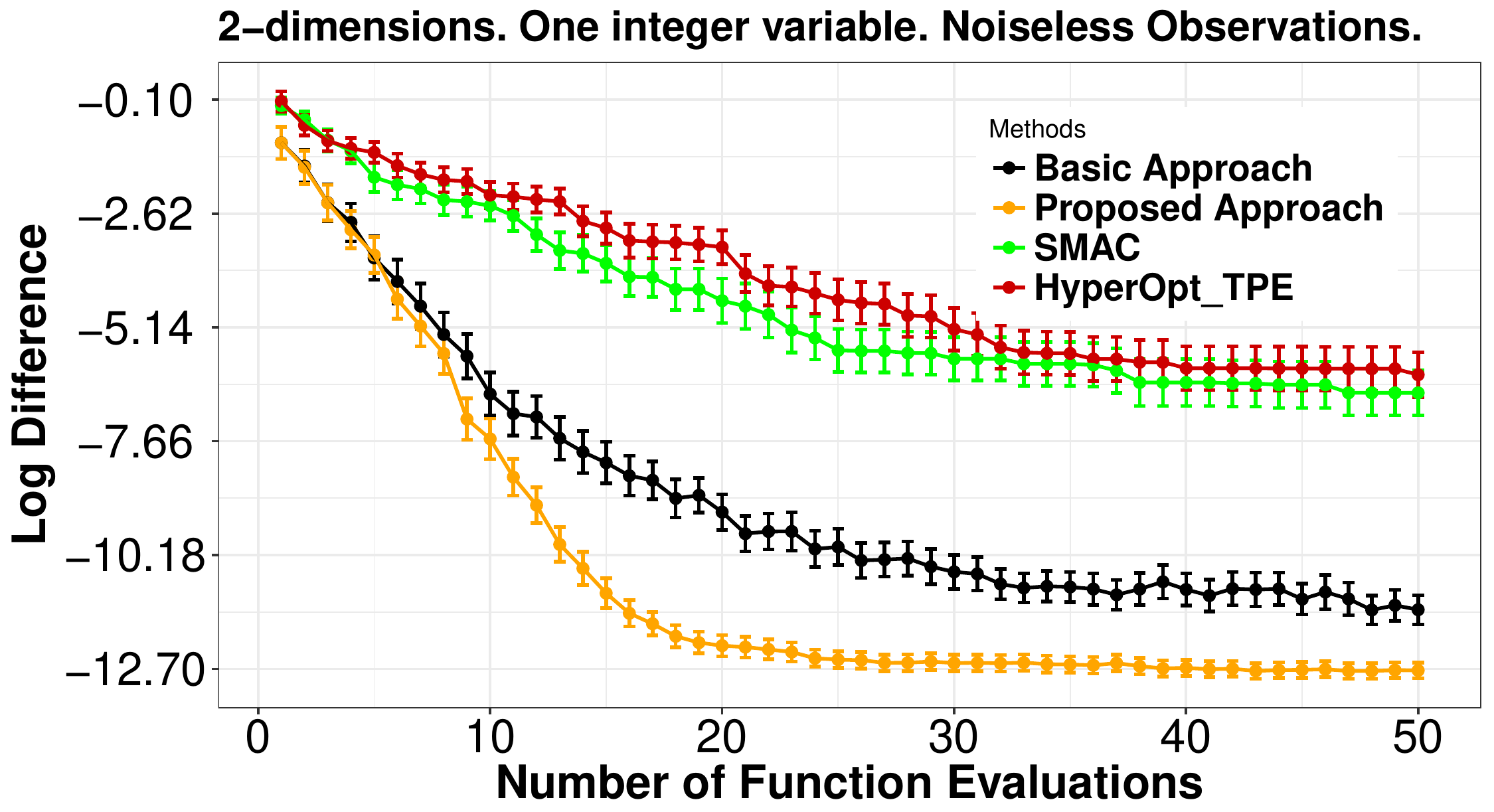} &
	\includegraphics[width=0.475\linewidth]{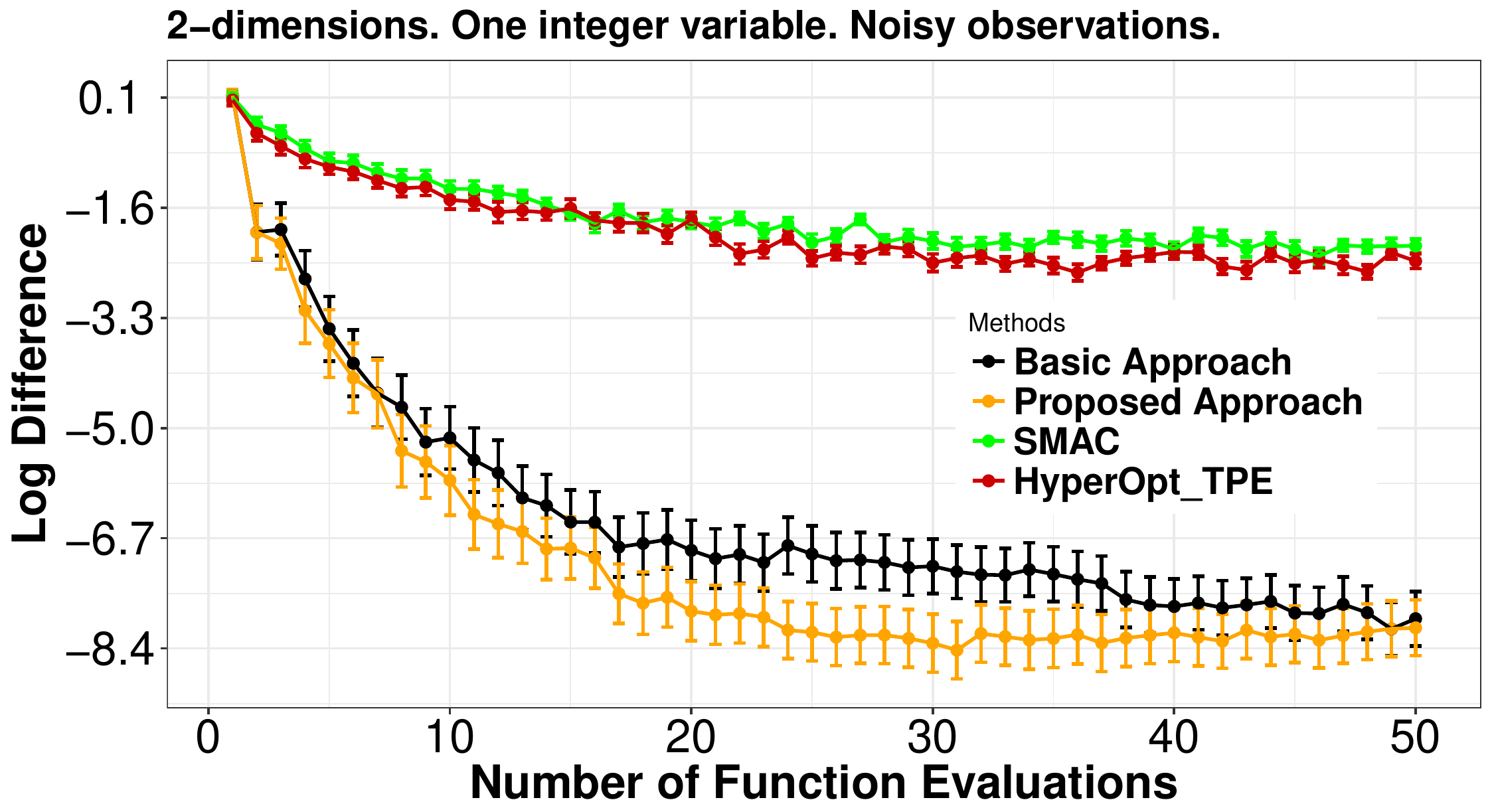} \\
        \includegraphics[width=0.475\linewidth]{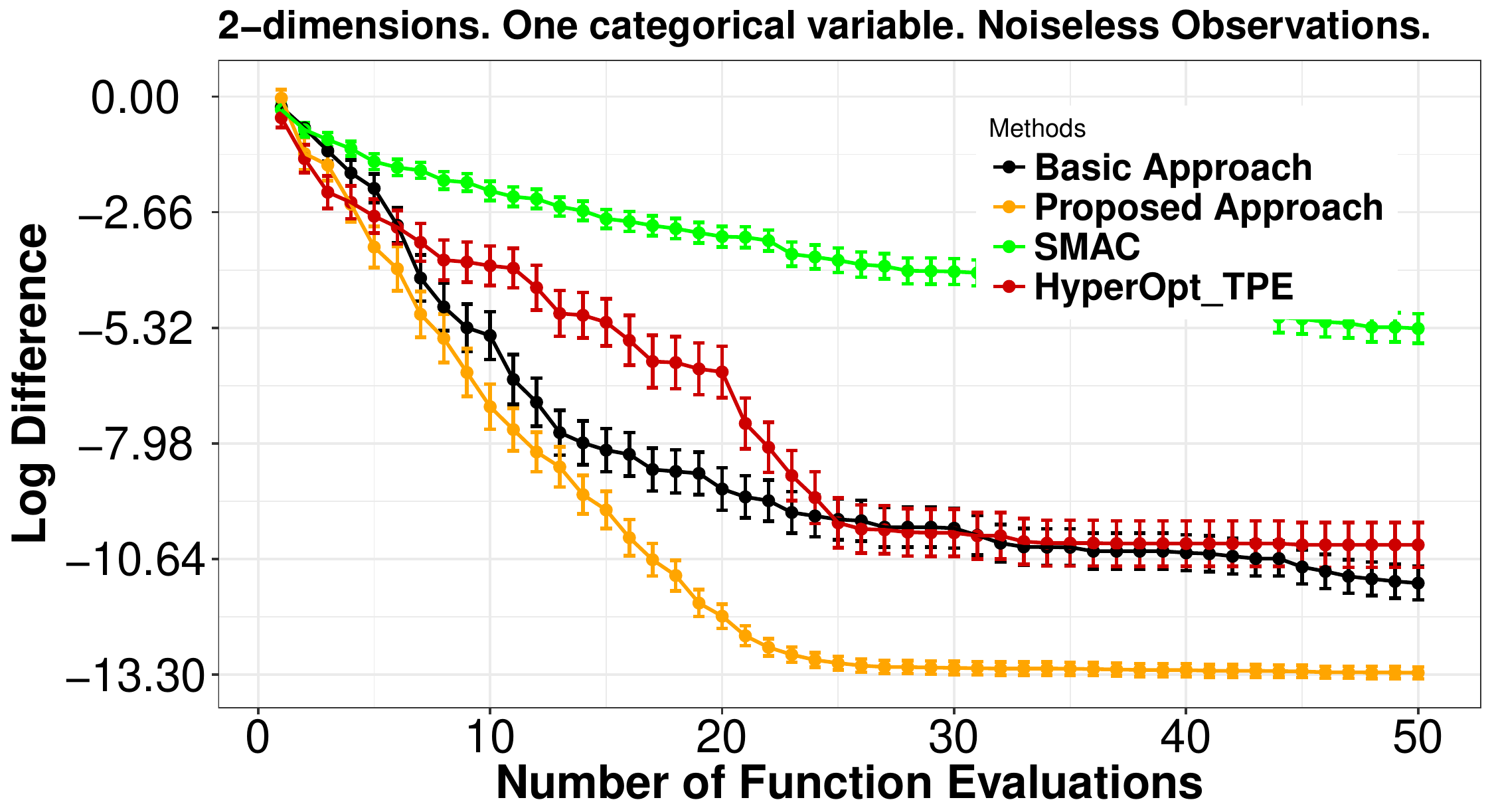} & 
        \includegraphics[width=0.475\linewidth]{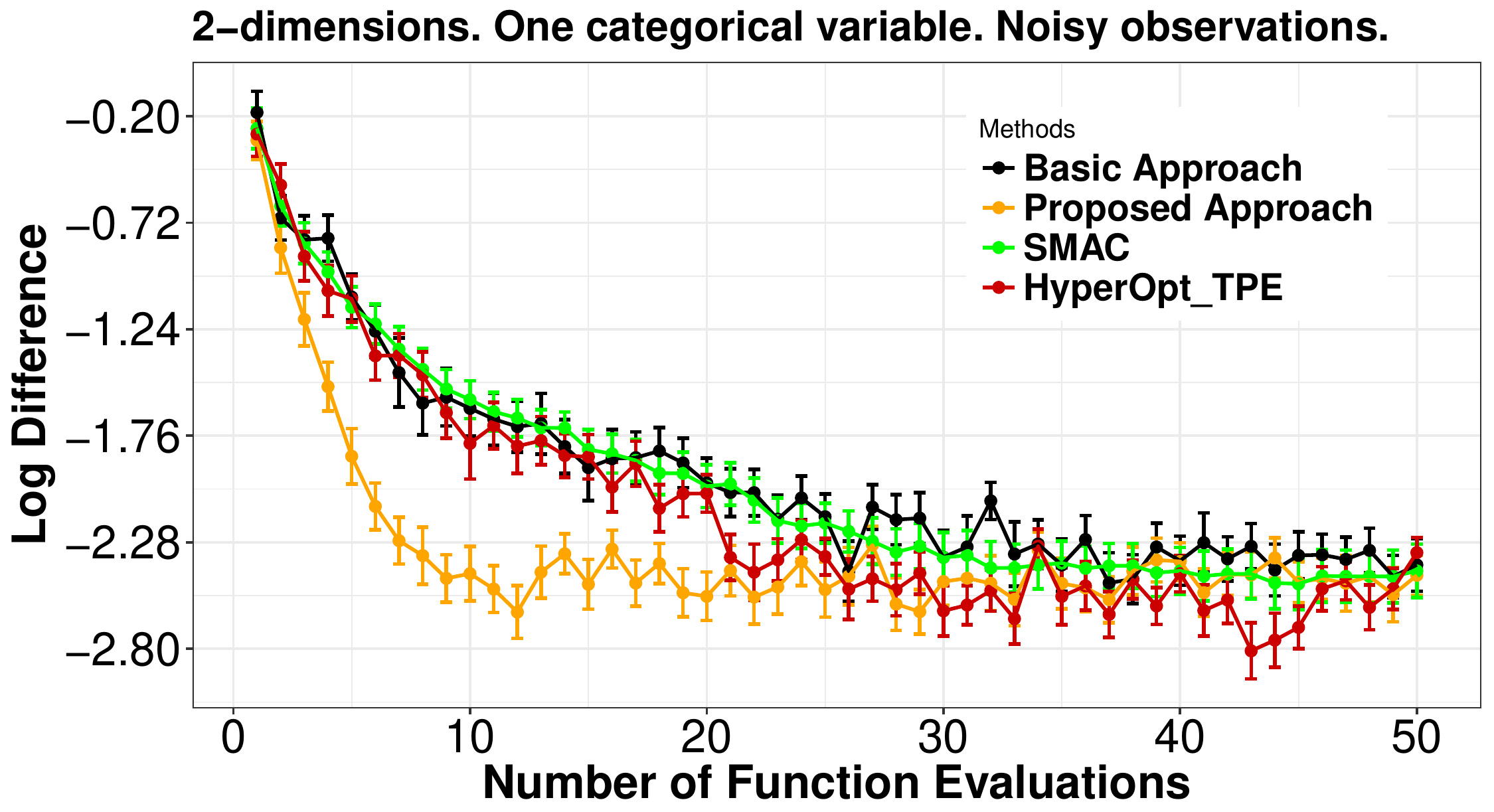} 
\end{tabular}
\caption{{\small Average results on the synthetic experiments with 2 dimensions. The first row
	shows results for a real and an integer-valued input variable. The second row shows results
	for a real and a categorical input variable. Left column is for noiseless observations. Right column
	is for noisy observations.}}
\label{fig:results_synthetic_2d}
\end{figure}

A first set of experiments considers $2$ input variables. In the integer case, the first variable is real
and the second variable can take $5$ different integer values. In the categorical case, the first variable is real
and the second is categorical, taking $5$ different values. Figure \ref{fig:results_synthetic_2d} shows 
the average results obtained by each method. This figure shows, as a function of the evaluations made, 
the average difference in absolute value between the objective associated to the recommendation made and the 
minimum value of the objective, in a log scale. We observe that the proposed approach significantly outperforms the basic 
approach and all other methods. More precisely, it is able to provide recommendations that are more 
accurate with a smaller number of evaluations, both for the noise and noiseless scenario. 
TPE outperforms SMAC and also the basic approach in the noisy setting. The bad results of TPE and SMAC are 
probably due to the use of an underlying model that is different from a GP. Therefore, they can suffer from model bias.

\begin{figure}[htb]
\begin{tabular}{cc}
        \includegraphics[width=0.475\linewidth]{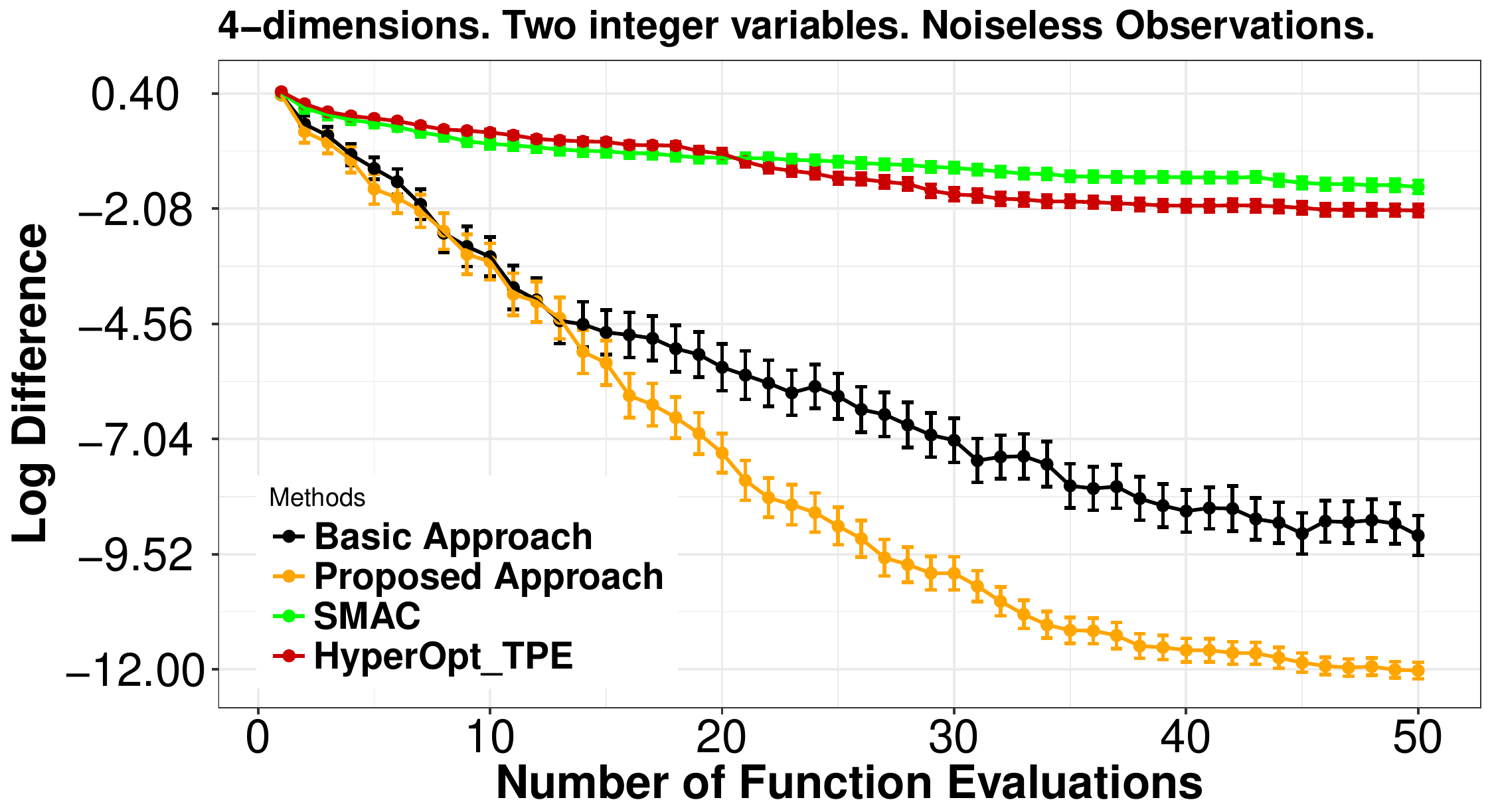}  &
        \includegraphics[width=0.475\linewidth]{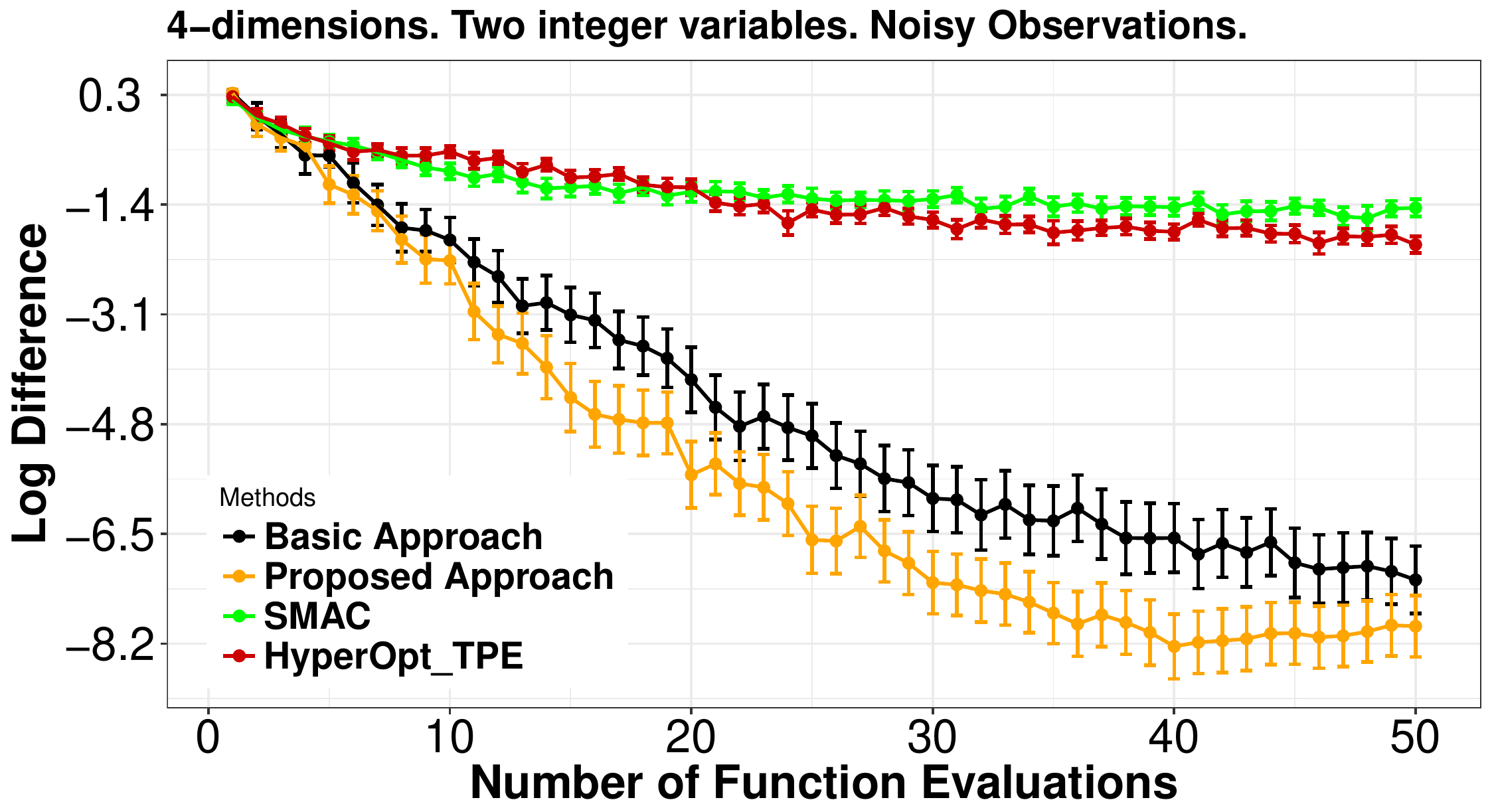} \\
        \includegraphics[width=0.475\linewidth]{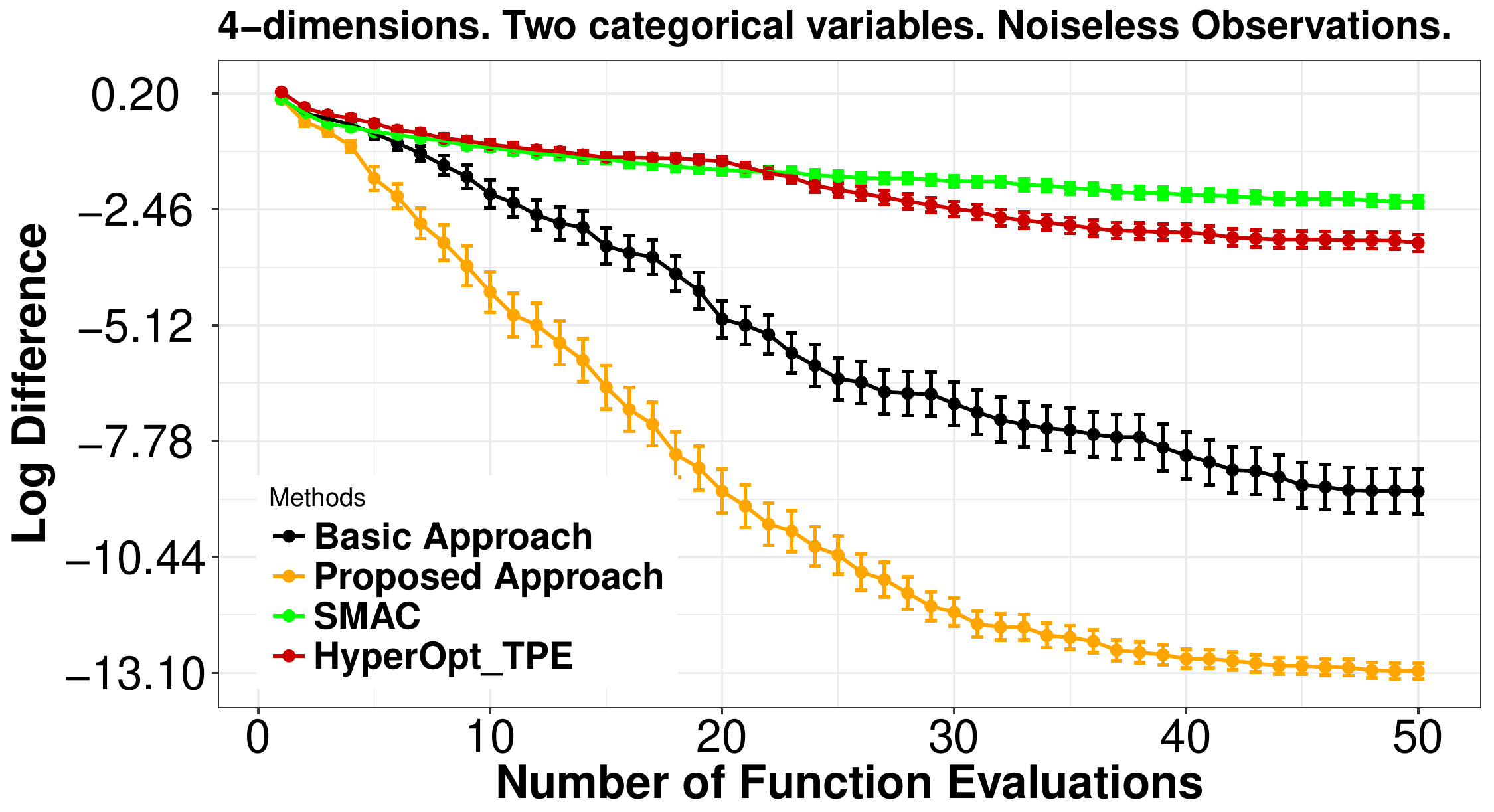}  &
        \includegraphics[width=0.475\linewidth]{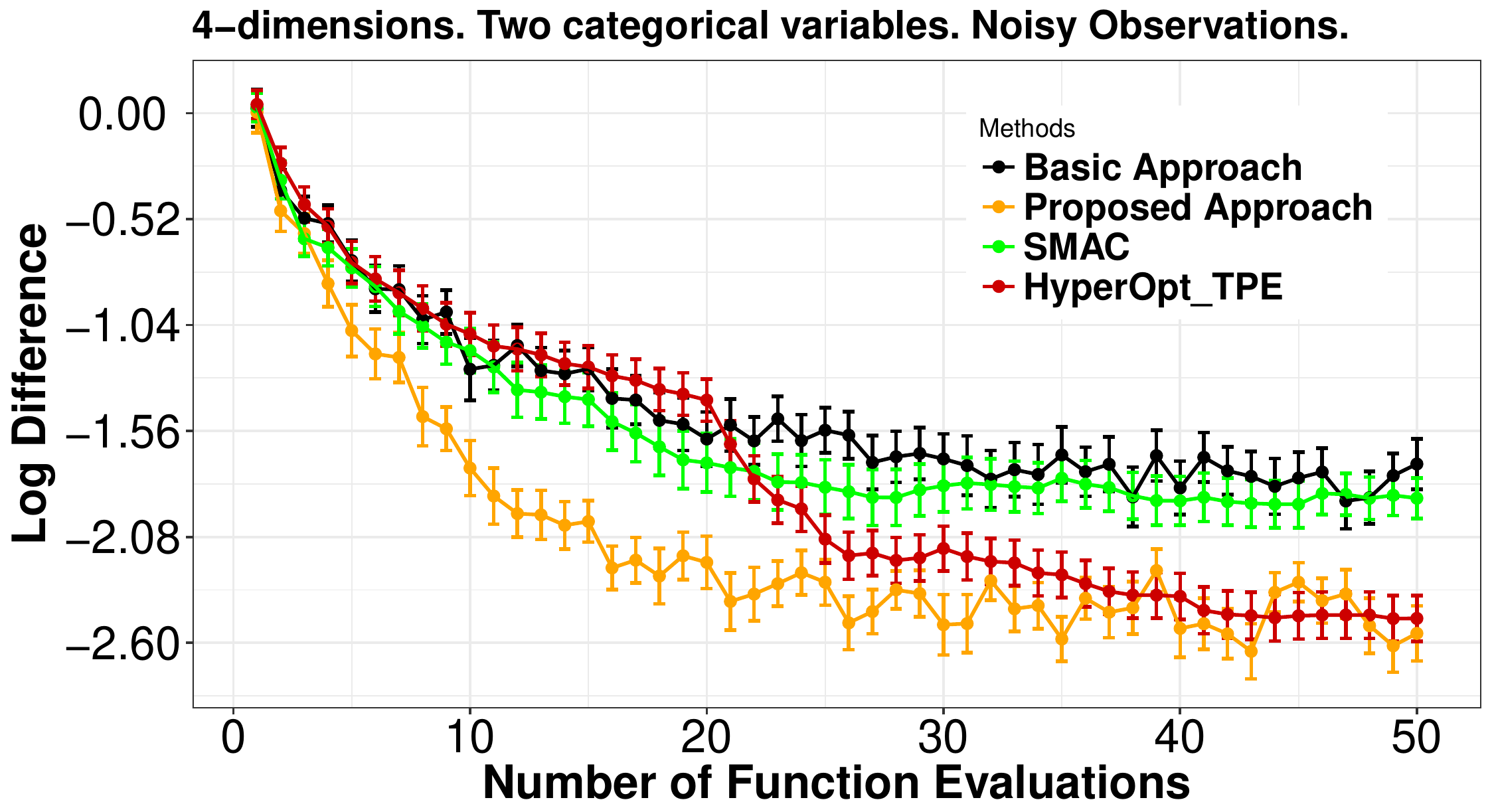} 
\end{tabular}
\caption{{\small Average results on the synthetic experiments with 4 dimensions. The first row
	shows results for two real and two integer-valued input variables. The second row shows results
	for two real and two categorical input variables. Left column is for noiseless observations. Right 
	column is for noisy observations.}}
\label{fig:results_synthetic_4d}
\end{figure}

A second set of experiments considers $4$ input variables. In the integer case, the two variables take real 
values, and two variables take $5$ different integer values. In the categorical case, two variables are real 
and two variables are categorical with $3$ different values each. Figure \ref{fig:results_synthetic_4d} 
shows the average results obtained by each method. Again, we observe that the proposed approach significantly 
outperforms the basic approach and all other methods. If we compare Figure \ref{fig:results_synthetic_2d}
and \ref{fig:results_synthetic_4d}, we observe that in this higher dimensional input space
the proposed approach gives even better results than in the previous setting. This makes sense, because when 
the number of dimensions is higher, the optimization problem is more difficult, and a better model 
can give much better results.

The functions that we have optimized in this section are extracted from a GP prior. Therefore, one can argue that 
the observed disadvantage of TPE and SMAC when compared to the proposed approach is in fact due to model bias,
as the GP matches the objective function in this case. In order to compare results
when there is model bias in the proposed approach, we carry out two real-world experiments in the next section.

\subsection{Real-world Experiments}

In this section, we compare all methods on the practical problem of finding the optimal 
parameters of a gradient boosting ensemble \citep{friedman2001greedy} on the digits dataset.
This dataset has 1,797 data instances, 10 class labels and 64 dimensions. It has been
extracted from the python package scikit-learn \citep{scikit-learn}.
Similarly, we also consider finding the optimal hyper-parameters of a deep neural network on 
the MNIST dataset \citep{lecun1998mnist}. This dataset has $60,000$ data instances, $768$ dimensions
and $10$ class labels. In this set of experiments, we use Predictive Entropy Search (PES) as 
the acquisition function, for both the basic and the proposed approach.

In the task of finding an optimal ensemble on the digits dataset, the objective that is considered for 
optimization is the average test log likelihood of the ensemble. This objective is evaluated using a 10-fold cross-validation 
procedure. Note that model bias can be an issue for all methods in this case, since the actual objective is unknown.
We consider a total of $125$ evaluations of the objective. 
A summary of the parameters optimized, their type and their range is displayed on Table \ref{table:1}. These 
parameters are: The logarithm of the learning rate, the maximum depth of the generated trees and the minimum number
of samples used to split a node in the tree building process. Importantly, while the first parameter can take 
real values, the other two can only take integer values.

\begin{table}[htb]
\centering
\caption{Names, types and range of the parameters optimized for the ensemble of trees.}
\begin{tabular}{ l | c | c }
 \hline
 {\bf Name} & {\bf Type} & {\bf Range} \\
 \hline
 Log Learning Rate & Real & $[-10,0]$ \\
 Maximum Tree Depth & Integer & $[1,6]$ \\
 Minimum Number of Samples to Split & Integer & $[2,6]$ \\
 \hline
\end{tabular}
\label{table:1}
\end{table}

\begin{figure}[htb]
        \begin{center}
	\begin{tabular}{c}
        \includegraphics[width=0.8\linewidth]{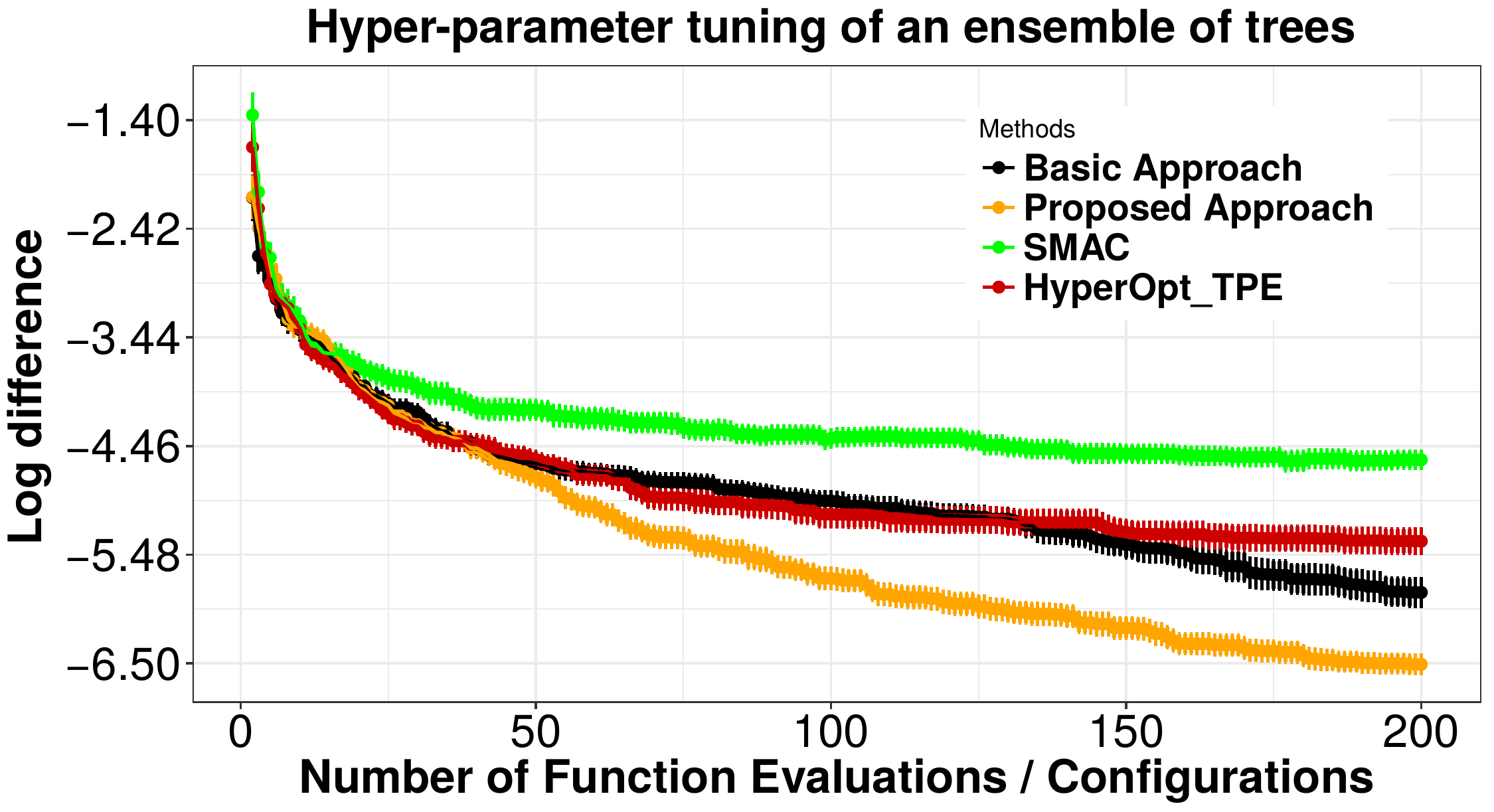} \\
        \includegraphics[width=0.8\linewidth]{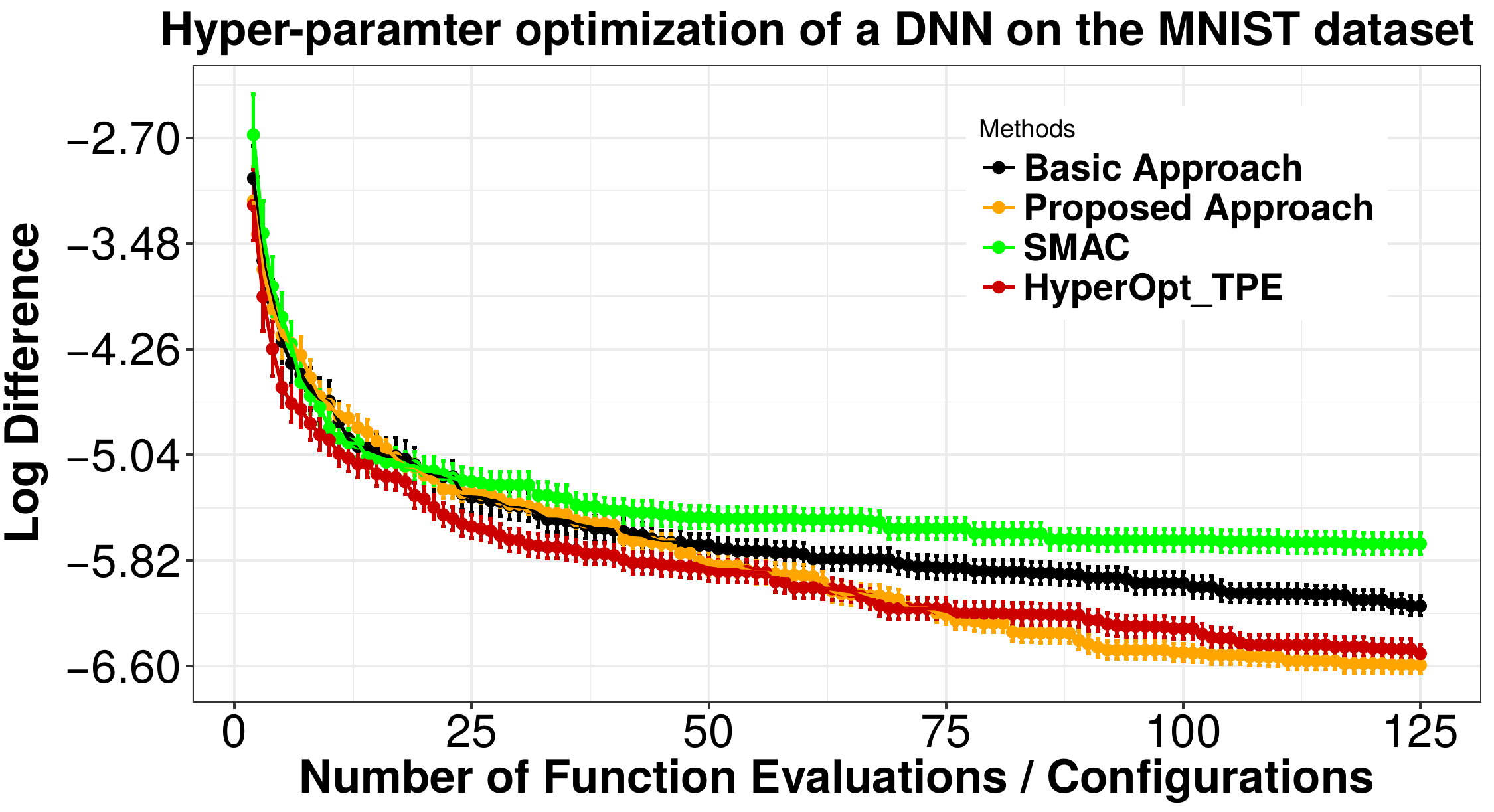} 
	\end{tabular}
        \end{center}
\caption{{\small (top) Average results on the task of finding an optimal gradient boosting ensemble on the digits dataset.
(bottom) Average results on the task of finding an optimal deep neural network on the MNIST dataset.}}
\label{fig:results}
\end{figure}

In each repetition of the experiment described (there are 100 repetitions) we consider a different 10-fold 
cross validation split of the data. The average results obtained are displayed in Figure \ref{fig:results} 
(top). This figure shows the average difference, in absolute value, between the test log-likelihood of the 
recommendation made and the best observed test-log likelihood, for that particular split, in a log scale. 
We observe that the proposed approach significantly outperforms the basic approach. More precisely, it is able
to find parameter values that lead to a gradient boosting ensemble with a better test log likelihood,
using a smaller number of evaluations of the objective. Furthermore, the proposed approach also performs
better than SMAC or TPE.

In the task of finding an optimal deep neural network on the MNIST dataset, the objective considered is 
the test log-likelihood of the network on a validation set of $10,000$ instances, extracted from the training set.
We consider $150$ evaluations of the objective. A summary of the parameters optimized, 
their type and their range is displayed on Table \ref{table:2}. These parameters are: 
The logarithm of the learning rate, the activation function and the number of hidden 
layers. The first parameter can take values in the real line. The second and third 
parameters are categorical and integer-valued. The number of units in each layer, is set
equal to $75$.

The average results obtained are displayed in Figure \ref{fig:results} (bottom). 
The figure shows the average difference, in absolute value, between the test log-likelihood 
of the recommendation made and the best observed test-log likelihood, in a log scale. 
Again, the proposed approach significantly outperforms the basic approach. More precisely, it is able
to find parameter values that lead to a deep neural network with a better test log likelihood
on the validation set, using a smaller number of evaluations of the objective. The proposed approach
also outperforms SMAC. However, TPE is only slightly worse than the proposed approach at the end 
and it outperforms the basic approach. We believe that the better results of TPE obtained in this 
problem can be a consequence of model bias in the GP that is used to fit the objective in the 
proposed approach.

\begin{table}[htb]
\centering
\caption{Name, type and range of the deep neural network parameters optimized.}
\begin{tabular}{ c | c | c }
 \hline
 {\bf Name} & {\bf Type} & {\bf Range} \\
 \hline
 Log Learning Rate & Real & $[-10,0]$ \\
 Activation Function & Categorical & Linear, Sigmoid, Tanh or ReLU \\
 Number of hidden layers & Integer & $[1,3]$ \\
 \hline
\end{tabular}
\label{table:2}
\end{table}

\section{Conclusions} \label{sec:conclusions}

BO methods rely on a probabilistic model of the objective function, typically a Gaussian process (GP), 
upon which an acquisition function is built. The acquisition function is used to select candidate
points, on which the objective should be evaluated, to solve the optimization problem
in the smallest number of evaluations. Nevertheless, GPs assume continuous input variables. 
When this is not the case and some of the input variables take categorical or integer values, 
one has to introduce extra approximations. 
A common approach before doing the evaluation of the objective is to use a one-hot encoding 
approximation for categorical variables, or to round the value to the closest integer, in the case 
of integer-valued variables. We have shown that this can lead to problems
as the BO method can get stuck, always trying to evaluate the same candidate point. 

The problem described is a consequence of a mismatch between the regions of the input space that have high acquisition
values and the points on which the objective is evaluated. A simple way of avoiding this problem 
is to do the approximations (a one-hot encoding in the case of categorical
variables or the approximation to the closest integer in the case of integer-valued variables) inside the
wrapper that is used to evaluate the objective. This technique works in practice, but it has the limitation
that it makes the objective constant in those regions of the input space that lead to the same configuration.
This constant behavior cannot be modeled by standard GPs.

In this paper we have proposed to modify the covariance function of the underlying GPs model to account for
those regions of the input space in which the objective should be constant. The transformation simply 
rounds integer-valued variables to the closest integer. In the case of categorical variables in which
one-hot encoding has been used, we simply set the largest extra variable equal to one and all the other equal to zero.
The consequence of this transformation is that the distance between those points of the input space 
that lead to the same configuration becomes zero. This enforces maximum
correlation between the GP values at those input values, leading to a constant behavior.

The proposed approach has been compared to a basic approach for dealing with categorical and integer-valued
input variables in the context BO and GPs. Furthermore, we have also compared results with two other approaches 
that can be used to solve these optimization problems and that can naturally account for integer and categorical
variables. Namely, SMAC and TPE. Several experiments involving synthetic and real-world experiments illustrate the 
benefits of the proposed approach. In particular, it outperforms the basic approach and SMAC and is most of the
times better or at least equivalent to TPE.

\section*{Acknowledgments}

The authors acknowledge the use of the facilities of Centro
de Computaci\'on Cient\'ifica (CCC) at Universidad Aut\'onoma de
Madrid. The authors also acknowledge financial support from the Spanish
Plan Nacional I+D+i, Grants TIN2016-76406-P, TEC2016-81900-REDT 
(MINECO/FEDER EU), and from Comunidad de Madrid, Grant S2013/ICE-2845.

\bibliography{jmlr_paper}
\bibliographystyle{apalike}

\end{document}